\documentclass[runningheads]{llncs}

\usepackage{eccv}

\usepackage{eccvabbrv}

\usepackage{graphicx}
\usepackage{booktabs}

\usepackage[accsupp]{axessibility}  

\usepackage{amsmath}
\usepackage{mathtools}
\usepackage{upgreek}
\usepackage{float}

\usepackage{hyperref}

\usepackage{orcidlink}

\begin{document}

\title{3DFlowRenderer: One-shot Face Re-enactment via Dense 3D Facial Flow Estimation} 

\titlerunning{3DFlowRenderer}

\author{Siddharth Nijhawan \and
Takuya Yashima \and
Tamaki Kojima}

\authorrunning{S.~Nijhawan et al.}

\institute{Sony Group Corporation}

\maketitle

\begin{abstract}
  Performing facial expression transfer under one-shot setting has been increasing in popularity among research community with a focus on precise control of expressions. Existing techniques showcase compelling results in perceiving expressions, but they lack robustness with extreme head poses. They also struggle to accurately reconstruct background details, thus hindering the realism.  In this paper, we propose a novel warping technology which integrates the advantages of both 2D and 3D methods to achieve robust face re-enactment. We generate dense 3D facial flow fields in feature space to warp an input image based on target expressions without depth information. This enables explicit 3D geometric control for re-enacting misaligned source and target faces.
  We regularize the motion estimation capability of the 3D flow prediction network through proposed "Cyclic warp loss" by converting warped 3D features back into 2D RGB space. To ensure the generation of finer facial region with natural-background, our framework only renders the facial foreground region first and learns to inpaint the blank area which needs to be filled due to source face translation, thus reconstructing the detailed background without any unwanted pixel motion. Extensive evaluation reveals that our method outperforms state-of-the-art techniques in rendering artifact-free facial images. 
\end{abstract}

\section{Introduction}
\label{sec:intro}
\begin{figure}[t]
  \centering
  \includegraphics[width=0.90\linewidth]{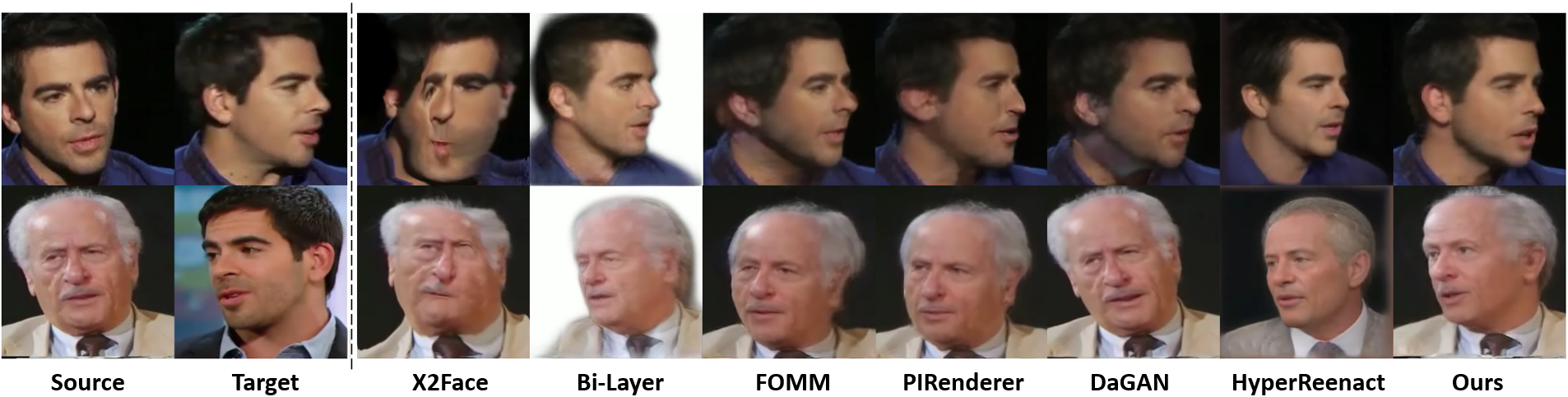}
  \vspace{-5pt}
  \caption{
  Qualitative comparisons with state-of-the-art methods in cases of extreme poses. Our method can perform well in both self identity re-enactment (first row) and cross identity re-enactment (second row), specially in cases where target pose is significantly different from source pose.
  }
  \label{fig:fig1_misalignment_comparison}
\end{figure}

One-shot facial re-enactment is the task of animating a facial expression of a single still source portrait image, based on the expressions in a target video. They have applications in various scenarios like online video conferencing, film making, social media, and virtual reality. However, it presents a highly challenging task, requiring the algorithm to dissect the 2D or 3D geometric structure of the input faces and perform animation while matching the expressions closely. At the same time, they must ensure realism in the generated faces.
Initial works like \cite{x2face} and \cite{bilayer} generated warping fields based on embeddings learned from input faces and used these fields to spatially transform the source image. Even though such methods display a good semantic control over facial expressions, they suffer from unwanted distortions when dealing with target poses which are significantly different from the source image (see \cref{fig:fig1_misalignment_comparison}). Further, warping field estimation networks in such methods are usually constrained only with style-based and perceptual losses which intuitively seems insufficient as the warped image should not only look realistic, but also display appropriate pixel-wise warping based on target expressions.

Instead of relying on warping operation, works like \cite{hyperreenact, oorloff2023oneshot, styleavatar} exploit the powerful image translation performance of StyleGAN \cite{stylegan} by manipulating its latent space through 2D or 3D facial priors. They achieve reasonable performance in editing facial expressions even in misaligned poses but have several drawbacks. The background information is either completely lost or changed in the final output. Moreover, they suffer from inaccurate pupil movement and hair texture sticking to the background. Moving away from rendering in 2D space, several works focus on representing 2D input as a triplane formulated volume which encodes the desired target motion in 3D space \cite{otavatar, hidenerf, latentavatar}. These methods employ volume rendering to generate multi-view consistent 3D head avatars. Such methods can successfully reconstruct human head avatars with high-fidelity rendering of fine facial textures. However, when generating talking heads, they suffer from rigid movements of head and torso, leading to a mismatch in motion between the head and the background.

In this paper, we propose a novel technique which blends the key aspects of both 2D and 3D methodologies to tackle the challenges outlined above. Firstly, we separate out the foreground and the background from source image to preserve its intrinsic details. The foreground includes the facial region and background  includes all the rest of the image. To enhance the robustness towards misaligned poses, we introduce a 3D warping block which transforms the source foreground in 3D feature space based on the target expression.
We describe the target expressions to be transferred by mapping its 3DMM parameters to a canonical latent space, which are then injected into the source facial statistics via 3D warping network, generating 3D facial flow across 3 dimensions: height, width and facial depth. We also design a 3D feature encoder tasked with extracting 3D appearance feature volume from 2D input source foreground. This feature volume is then warped using the 3D facial flow, generating a set of warped features with explicit 3D information. Learning the facial flow and performing warp operation in 3D ensures that our network can spatially transform the source face while considering its 3D geometry, specifically in highly misaligned cases. Further, a 3D feature decoder converts the set of warped feature maps back to a RGB image, containing individual in source image with target expressions. This unique blend of 3D warping with 2D image generation allows us to regularize the motion estimation capability of the 3D warping network through proposed "Cyclic warp loss", which otherwise would have been limited to constraining the perceptual quality only, leading to inaccuracies in expression rendering.

To enhance the naturalism of foreground obtained from 3D warping network, we propose a block titled ‘TransUNet’ which is based on UNet \cite{unet} architecture and performs image refinement. It removes unwanted artifacts from the warped face and generates facial textures which are not visible in the original source image due to occlusion. We project this refined foreground back to the extracted background, and obtain an image with several blank spaces due to the source face translation. To generate the final output, we again use the TransUNet block with different weights to perform image-to-image translation, which inpaints the blank spaces based on neighboring background pixels’ information. Since our TransUNet block is used for both image-to-image transfer (refinement) and translation (inpainting) tasks, we call it "TransUNet". Thus, our work ensures that background information is preserved well without any loss or unwanted motion with facial foreground. In total, we combine the capability of 2D methods to offer direct control over modifying facial features with non-rigid movements and multi-view consistency of 3D methods together, to realize face re-enactment under one-shot setting.

Our key contributions can be summarized as follows:
\begin{itemize}
\setlength{\parskip}{0cm}
\setlength{\itemsep}{0cm}
\item We propose a network for learning dense 3D facial flow to warp source image in 3D feature space based on target expressions using only 2D images and regularize its flow estimation capability in 2D space through proposed Cyclic warp loss. To the best of our knowledge, this is the first work integrating the advantages of both 2D and 3D methods to achieve robust face re-enactment using 3DMM-based priors.
\item We show the effectiveness of rendering facial foreground and background separately; allowing reconstruction of high-frequency details in the background for added realism.
\item We conduct extensive evaluation on VoxCeleb \cite{voxceleb} dataset, both quantitatively and qualitatively, showcasing realistic rendering capability of our network while preventing leakage (or motion) and loss of background pixels. It is also robust to extreme cases of head pose changes and shifts.
\end{itemize}

\section{Related Work}
\label{sec:relatedwork}
We categorize the existing works across 3 dimensions: image warping based, StyleGAN based, and volumetric 3D human head reconstruction.

\textbf{Image warping based.} These methods focus on representing the facial motion to be transferred using deformations (or flow fields) and use these to spatially transform the input source image. Wiles et al. \cite{x2face} was one of the first works to employ warping directly on source image and perform re-enactment. \cite{bilayer, dpe, animatingarbit, fomm} applied the same concept to explicitly predict warping flow, however, leveraged the sparse priors defined by 2D facial landmarks to map target motion. To interpolate the flow between source and target frame, Siarohin et al. \cite{animatingarbit} proposes usage of local 2D affine transformations. \cite{x2face, facegan, zhao2021sparse} represent the target facial motion to be transferred using a predefined set of landmarks. However, due to the limitation of 2D priors in appropriately perceiving the target facial expressions and deformations with changes in emotions, several works shifted to 3D facial priors. Wang et al. \cite{facevid2vid} generates 3D canonical key points from the input faces and warps them in 3D space. Hong et al. \cite{dagan} estimates pixel-wise depth maps to generate a set of sparse facial keypoints for warping the input image. But the expressive capability of such models weakens in cases with large differences between source and target expressions due to sparsity of the learned key points. Some methods leverage the power of 3D morphable models to parametrically map the target facial motion into a canonical space and use them as 3D priors. \cite{pirenderer, headgan} predict the warping flow by computing 3DMM parameters which allows explicit control of input face. Since, the warping is limited to 2D pixel-space, these methods fail to render realistic looking images in extreme pose variations. We tackle this challenge by learning dense 3D facial flow and spatially warping source in 3D feature space, driven by 3DMM-based priors instead. Our method falls under the category of image warping as such techniques offer direct control over modifying facial features with flexibility in motion. 

\textbf{StyleGAN based.} Mapping the input images into the latent space of a pre-trained StyleGAN2 \cite{stylegan2} has been a popular choice among several recent works \cite{stylesync, hyperreenact} because of its ability to generate high-quality portrait images. \cite{stylerig, interfacegan} try to edit source images by computing different semantics through latent space manipulation in StyleGAN2 generator. Kang et al. \cite{megafr} obtains deformation between source and target image through 3D priors and encodes it as a  $\mathcal{W}^+$ space offset which is added to the existing latent space of StyleGAN2.  Some methods use StyleGAN as a refiner to generate artifact free outputs. Yin et al. \cite{styleheat} obtains a set of latent codes and feature maps through GAN inversion, warps them based on target motion, and generates final image through StyleGAN. Wang et al. \cite{styleavatar} uses a pair of StyleGAN-based networks to generate feature maps from input 3D face mesh and output the final image, sequentially. Though these methods are powerful in generating high-quality portrait images with meaningful control over facial poses and shapes, rendering fine-grained expressions like pupil movement, blinks, etc. is a challenge for them. Moreover, since they rely heavily on a pretrained StyleGAN, output suffers from hair texture sticking issues and loss of background information. 

\textbf{Volumetric 3D human head reconstruction.} For improving multi-view consistency of generated portrait images, methods like \cite{fenerf, headnerf} utilize neural representations such as NeRF \cite{nerf} for editable and 3D-consistent human avatars. \cite{next3d, hidenerf} further enhance the controllability of animating facial expressions by integrating 3DMM parameters. Park et al. \cite{nerfies} represents the facial motion using several spatial points which are used as a condition for generating deformation field. Similarly, Gafni et al. \cite{dynamicnerf} synthesizes the facial image by volume rendering based on estimated 3DMM parameters of target face. \cite{otavatar,latentavatar,learningnerf} decompose the 2D input into triplane features which are injected with target facial motion in latent space and use volume rendering to generate multi-view consistent synthetic images. Volumetric methods perform well in reconstructing human head avatars, however, face an inherent limitation in capturing dynamic head or torso movements in the generated avatar, resulting in rigid outputs that diminish realism. Consequently, they are not a suitable choice for effectively transferring facial expressions from one individual to another. Therefore, we focus our attention on developing an image warping based methodology which blends the advantages of both 2D (warping) and 3D (volumetric) approaches to realize robust face re-enactment.

\begin{figure}[t]
  \centering
  \includegraphics[width=0.95\linewidth]{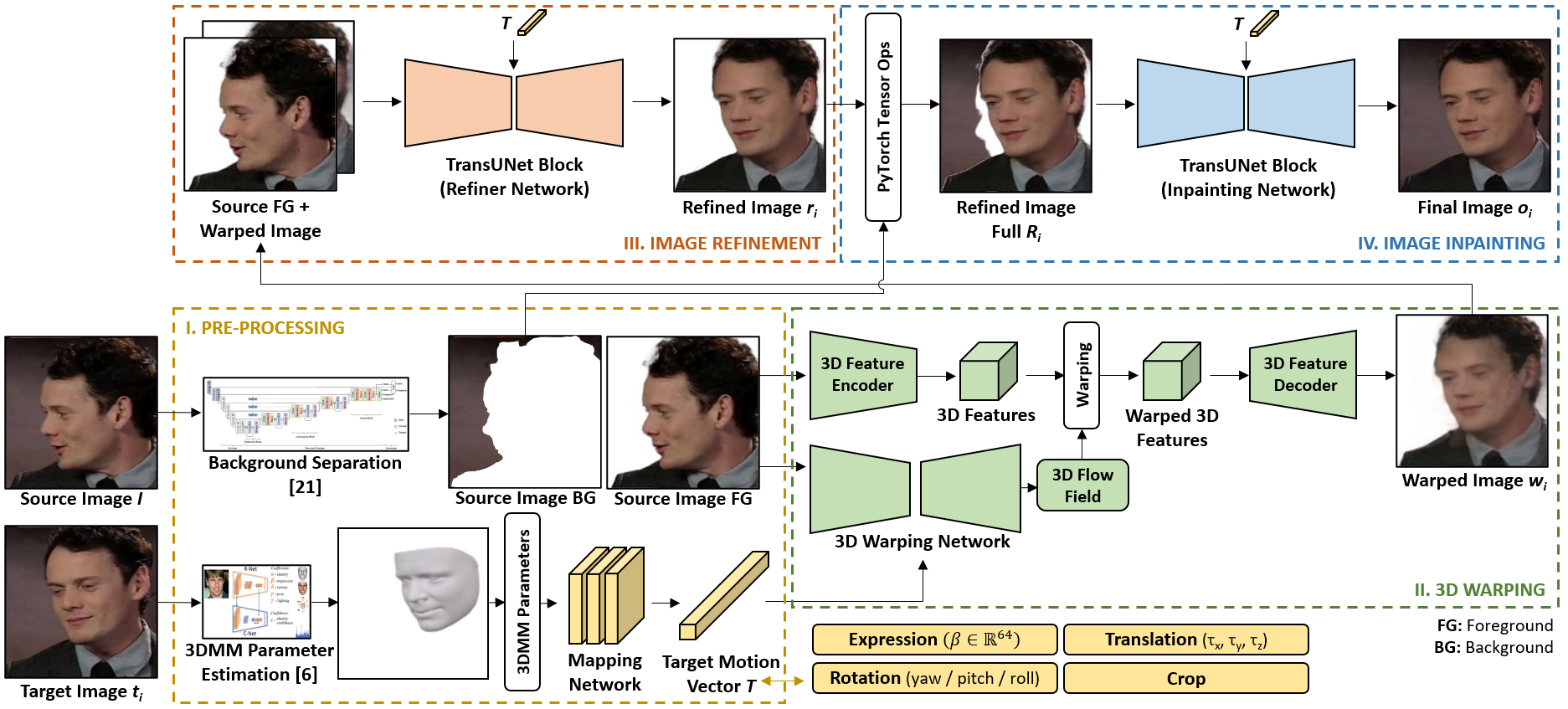}
  \vspace{-5pt}
  \caption{
  Overview of 3DFlowRenderer.
  %
  }
  \label{fig:fig2_methodology}
\end{figure}

\section{Methodology}
\label{sec:methodology}
For performing the task of one-shot face re-enactment, given a source image with a face $I$, and a video with target face having $N$ total frames, $\{t_1,t_2,...,t_N\}$, where $t_i$ is the ${i^{th}}$ video frame. Our method generates a video $\{o_1,o_2,...,o_N\}$ on a frame-by-frame basis, where each frame $o_i$ consists of the individual in $I$ matching the expressions and motions of corresponding frame $t_i$. 

The 4-stage architecture of our proposed method is illustrated in \cref{fig:fig2_methodology}. In the first stage (pre-processing, \cref{subsec:pre-processing}), we perform background separation of the source image and 3DMM parameter estimation. The second stage (3D warping, \cref{subsec:3dwarping}) computes the dense 3D facial flow field and warps the source foreground based on target expressions to generate a warped foreground. The third stage (image refinement, \cref{subsec:refinement}) refines this warped foreground and further transfers the motion of fine-grained facial regions. The final stage (image inpainting, \cref{subsec:inpainting}) blends the refined foreground with source background and inpaints the blank space to reconstruct the detailed final image.

\subsection{Pre-processing}
\label{subsec:pre-processing}
We perform 3 key tasks in the pre-processing stage: background separation in source image, 3DMM parameter estimation of target image and target motion vector generation. 

\textbf{Background separation.}  To accurately reconstruct the high-frequency details in the background and prevent the leakage of background pixels into the facial foreground region, we subtract the background from source image and feed only the facial foreground into the network for further computation. We choose to use the algorithm proposed in RVM \cite{rvm} due to its light-weight architecture and good performance in occluded scenarios.

\textbf{3DMM parameter estimation.} Similar to existing works \cite{pirenderer, headgan, styleheat}, we leverage 3DMM parameters as 3D priors for semantically meaningful control and modeling of target facial motion. 3DMM expresses the 3D shape $S$ of a human face in a controllable mathematical form:
\begin{equation}
  S = \hat{S} + \alpha{_{id}}{B_{id}} + \beta{_{exp}} {B_{exp}},
  \label{eq:3dmm}
\end{equation}
where $\hat{S}$ defines the average face shape. $B_{id}$ and $B_{exp}$ are bases of identity and expression, computed using 200 scans of human faces through Principal Component Analysis \cite{5279762}. The coefficients controlling facial shape and facial expressions are defined as $\alpha \in R^{80}$ and $\beta \in R^{64}$, respectively. Moreover, we can express the rotation of the head with parameters of yaw, pitch, and roll $R \in SO(3)$ and translation along $(x, y, z)$ axes as $\uptau \in R^3$. Alongside relative face translations, we use crop parameters $c \in R^3$ to express the absolute facial position. Therefore, we model the target facial motion with a parameter set containing 73 coefficients $p = \{\beta, R, \uptau, c\}$. We leverage 3D face reconstruction model proposed in Deep3DFaceRecon \cite{deep3dfacerecon} to extract these coefficients from real-world portrait images for both training and inference. $\hat{S}$ and $B_{id}$ is adopted from 2009 Basel Face Model \cite{bfm} and the expression bases $B_{exp}$ are built from Face-Warehouse \cite{facewarehouse}. To average out the errors accumulated during 3DMM parameter estimation, we pass a window of continuous frames $\{t_{i-k},...,t_{i},...t_{i+k}\}$ with radius $k$ to the estimation network and generate coefficients of these corresponding adjacent frames. Thus, we can prevent any degradation in performance due to mismatch between predicted coefficients and the actual target facial motion.

\textbf{Target motion vector generation.} We map the coefficients which describe target motion to a latent space producing a target motion vector $T$ for window of frames, $\{T_{i-k},...,T_{i},...T_{i+k}\}$. This is done by using a series of 1D convolutions and center crop operations. The motion vector is used to inject the target motion into each convolution layer via adaptive instance normalization (AdaIN) \cite{adain} in the later stages.

  


\subsection{3D Warping}
\label{subsec:3dwarping}
The previous methods \cite{pirenderer, styleheat, headgan} employ a warping network following an auto-encoder design to spatially transform a source image in 2D pixel space. Being confined within 2D is a limitation of the network's ability to control explicit 3D geometry of a human face. Thus, the output develops unwanted artifacts and distortions in cases where the target pose differs significantly from the source pose. To mitigate this issue, we introduce 3D warping. Existing works like Wang et al. \cite{facevid2vid} also perform warping in 3D feature space. They estimate a set of 3D facial keypoints from the input 2D images, and apply a series of deformations based on target facial orientation to generate a composited flow field. However, due to the sparsity of this flow field, they suffer from inaccurate expression transfer in cases with large differences between source and target expressions. Our 3D warping pipeline works on the same principle, however, computes a dense 3D facial flow field driven with 3DMM coefficients as prior, enhancing its overall expressive capability while being robust to large expression and pose variations.

We first dissect the RGB space into depth and channel information, and apply a series of encoder sub-blocks containing 3D convolutions followed by decoder sub-blocks to estimate flow field. To inject target motion into source image statistics, we employ AdaIN operation after every convolution block. Since feature maps have three spatial dimensions alongside channels, we modify the AdaIN blocks to incorporate facial depth by using 3D scale and bias parameters. A 3D feature encoder is used to extract 3D appearance features from 2D source image. Similar to 3D warping network's encoder, it projects 2D features into 3D and applies a series of 3D convolution blocks. The output feature volume is warped using the dense facial flow, generating a set of warped feature volume. To effectively train the motion estimation capability of 3D warping network, we use several loss functions described in \cref{subsec:training_strategies}. For this purpose, we use a 3D feature decoder to convert the warped feature volume back into 2D image. It merges the depth and channel space into RGB space and applies a series of decoder sub blocks, now with 2D convolutions, and up-sampling blocks to generate warped image.


\subsection{Image Refinement}
\label{subsec:refinement}
The generated warped image might contain several unwanted artifacts caused by the warping operation. Also, warping network does not generate contents which do not exist in the original source due the translation and occlusion of the face. To solve this issue, we propose a block based on UNet \cite{unet} design, titled "TransUNet". It takes source and warped images as input, and performs image-to-image transfer, refining the warped image. We choose UNet because of its capability to efficiently capture the features at both local and global level. Moreover, its segmentation-driven synthesis enhances the rendering quality of finer facial regions like pupils, eyes, teeth, lips, etc.  We add direct skip connections to preserve original source information and prevent texture loss after each down-sampling stage. We again inject target motion vector into input statistics via AdaIN operation after every convolution layer inside down-sampling blocks.

\subsection{Image Inpainting}
\label{subsec:inpainting}
Final stage involves generating the background with high-frequency details for added realism. First, we project the output of the image refinement network which contains only foreground information to the background of the source image, extracted during pre-processing stage. After the projection, some pixels could be blank due to translation of foreground, and hence can be treated as a classic inpainting problem. To reconstruct the detailed background, we perform the task of filling the blank space in correlation to neighboring pixels with another TransUNet block. This also reduces unwanted pixels leaking into the facial regions due to large pose changes. The output of this network will be the final image containing individual in source image with target facial expressions and reconstructed background.

\subsection{Training Strategies}
\label{subsec:training_strategies}
To achieve both intuitive control over facial expressions and better generalization, we train our networks in two phases. In the first phase, we pre-train the blocks performing 3D warping alongside the inpainting network. This ensures that 3D warping blocks are able to estimate meaningful deformations and good quality warped images aiding the convergence of refiner network's objective functions. Further, the same source-target image pairs can be utilized as the input and the ground-truth, respectively, for pre-training the inpainting network. This eliminates the need of preparing a separate dataset specifically for optimizing the inpainting network. We train the entire network together in the second phase. To compute losses with ground-truth during training, we use the same individual in the source image and the target video. However, our model can generalize well to different identities during inference, as shown in \cref{sec:evaluation}.

\begin{figure}[ht]
  \centering
  \includegraphics[width=0.80\linewidth]{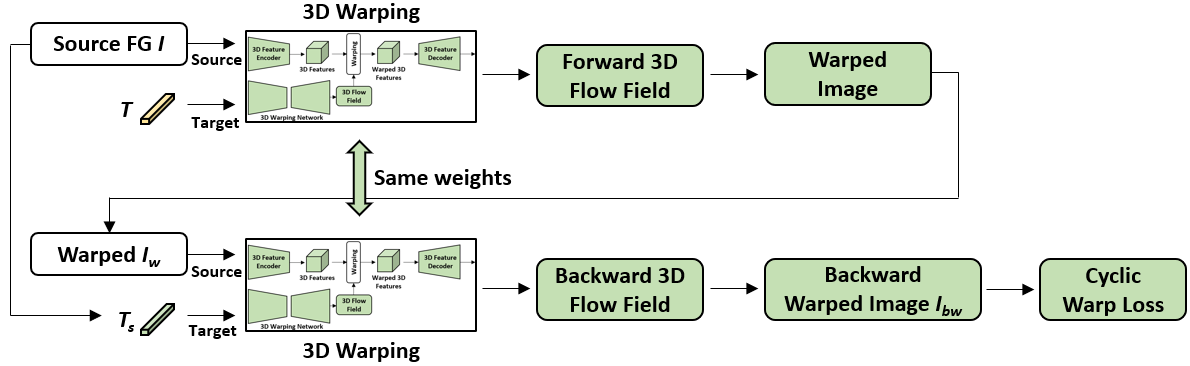}
  \vspace{-5pt}
  \caption{
  The proposed methodology to compute Cyclic warp loss $\mathcal{L}_{cw}$ during training.
  }
  \label{fig:fig6_cyclicwarploss}
\end{figure}

Similar to existing works \cite{pirenderer, styleheat}, 3D warping stage is constrained with a perceptual warp loss $\mathcal{L}_{pw}$ computed between the generated 2D warped image frame $w_i$, and the target frame $t_i$. It is based on \cite{perceptual_loss}, which computes $L1$ distance between activation maps of the pre-trained VGG-19 network \cite{vgg19}: 
\begin{equation}
  \mathcal{L}_{pw} = \sum_{j}||\phi^j(w_i) - \phi^j(t_i)||_1 ,
  \label{eq:perceptual_warp_loss}
\end{equation}
where $\phi^j(.)$ is the activation map of the $j^{th}$ layer in VGG-19 network. For the purpose of computing loss values, $t_i$ refers to the foreground of target frame. The minimization of perceptual loss ensures the perceptual similarity of generated images to ground-truth focusing on realism of output image. However, this is insufficient as it does not optimize the accuracy of dense flow prediction. Therefore, we propose another loss function titled "Cyclic warp loss", $\mathcal{L}_{cw}$, which regularizes the motion field estimation capability of 3D warping network in a cyclic manner. 

\cref{fig:fig6_cyclicwarploss} illustrates how we compute $\mathcal{L}_{cw}$ during training. First, we feed the source image $I$ and target motion vector $T$ for current frame to the 3D warping stage and generate a 2D warped image using the estimated forward facial flow. Then we feed this warped image back into the warping stage with same weights, but now as a new source, say, $I_w$. Further, we assume the expressions in $I$ as the target expressions for this stage, and compute motion vector $T_s$ accordingly. This is cyclic in the sense that warped image is now the source and original source is now the target. Intuitively, the 3D warping block should now estimate the motion field to deform $I_w$ back into $I$, termed as backward flow, generating a new backward-warped source image $I_{bw}$. Therefore, we can now constraint the $L1$ distance between newly generated $I_{bw}$ and original $I$:
\begin{equation}
  \mathcal{L}_{cw} = ||I_{bw} - I||_1 .
  \label{eq:cyclic_warp_loss}
\end{equation}

\begin{figure}[ht]
  \centering
  \begin{subfigure}{0.48\textwidth}
    \centering
    \includegraphics[width=0.9\linewidth]{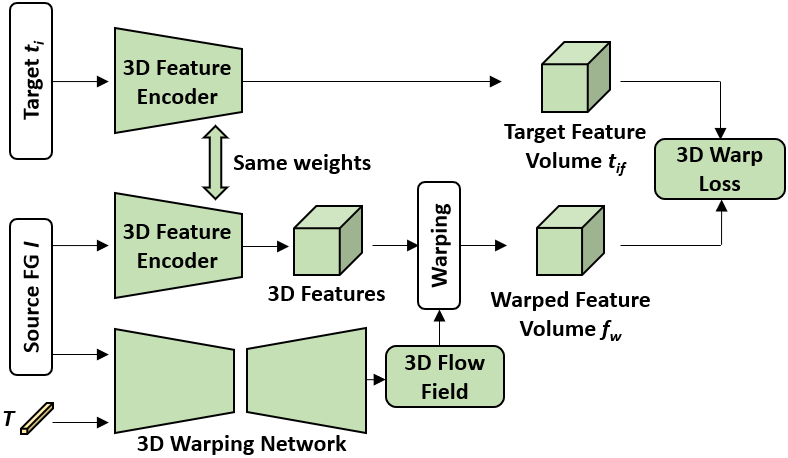}
    \caption{3D Warping Loss $\mathcal{L}_{3dw}$}
    \label{fig:fig7_a_3dwarploss}
  \end{subfigure}
    \hfill
  \begin{subfigure}{0.48\textwidth}
    \centering
    \includegraphics[width=0.9\linewidth]{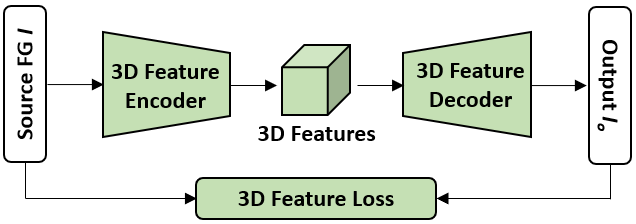}
    \caption{3D Feature Loss $\mathcal{L}_{3df}$}
    \label{fig:fig7_b_3dfeatureloss}
  \end{subfigure}
  \vspace{-5pt}
  \caption{The proposed methodology to compute (a) 3D warping loss $\mathcal{L}_{3dw}$ and (b) 3D feature loss $\mathcal{L}_{3df}$ during training.}
  \label{fig:fig7_3dlosses}
\end{figure}

The impact of Cyclic warp loss on rendering performance is investigated in \cref{subsec:ablation_study}. We also regularize each components of 3D warping stage. For 3D warping network, we use 3D warp loss $\mathcal{L}_{3dw}$ (\cref{fig:fig7_a_3dwarploss}) by computing mean-squared error (MSE) between warped feature volume $f_w$ and target feature volume $t_{if}$:
\begin{equation}
  \mathcal{L}_{3dw} = MSE(f_w - t_{if}).
  \label{eq:3d_warp_loss}
\end{equation}

$t_{if}$ is obtained by passing target frame $t_i$ through 3D feature encoder. We use the pipeline illustrated in \cref{fig:fig7_b_3dfeatureloss} to regularize 3D feature encoder-decoder pair. First, we obtain 3D feature volume for a given source frame $I$. Then, instead of applying warping, it is directly passed through 3D feature decoder to obtain the 2D output $I_o$. Intuitively, $I_o$ should be as close to $I$ as possible since no warping is applied. Therefore, we can compute 3D feature loss $\mathcal{L}_{3df}$ as the perceptual distance between $I_o$ and $I$:
\begin{equation}
  \mathcal{L}_{3df} = \sum_{j}||\phi^j(I_o) - \phi^j(I)||_1.
  \label{eq:3dfeature_loss}
\end{equation}

\begin{figure}[ht]
  \centering
  \includegraphics[width=0.7\linewidth]{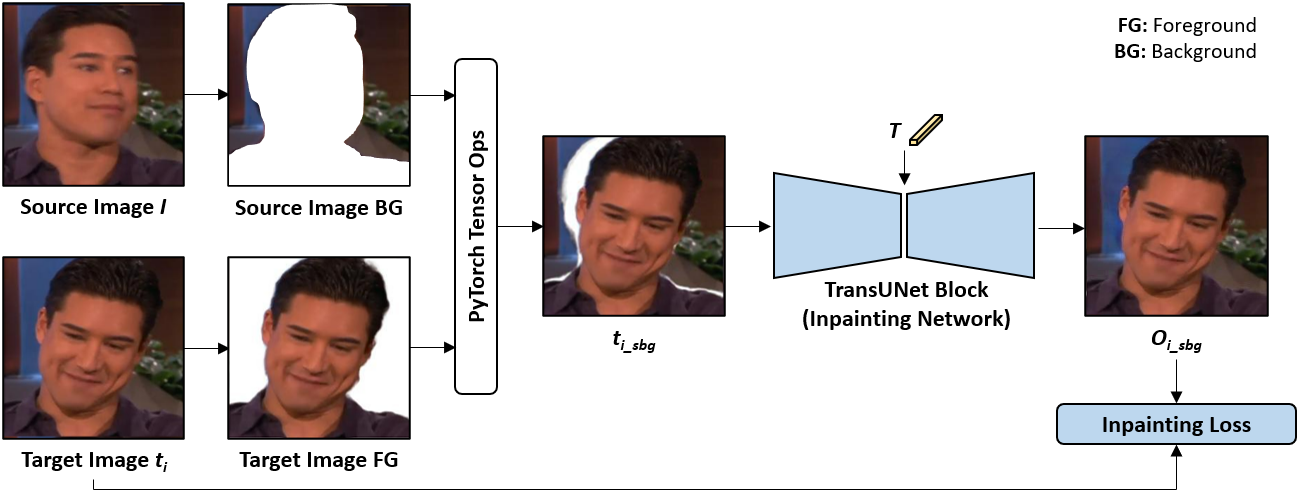}
  \caption{
  The proposed methodology for training the inpainting network in the first phase.
  }
  \label{fig:fig8_inpaintingloss}
\end{figure}

We train the inpainting network in two phases with inpainting loss $\mathcal{L}_{inp}$. During the first phase, we extract foreground from target frame $t_i$ and project it on to source background, producing an image $t_{i\_sbg}$ with several blank spaces (\cref{fig:fig8_inpaintingloss}). $t_{i\_sbg}$ is passed through the inpainting network, generating output $O_{i\_sbg}$. The ground-truth for this image should be the target frame $t_i$. Therefore, we can minimize perceptual loss between them,
\begin{equation}
  \mathcal{L}_{inp} = \sum_{j}||\phi^j(O_{i\_sbg}) - \phi^j(t_i)||_1.
  \label{eq:inpaintingloss}
\end{equation}

In the second phase, we can simply replace $t_{i\_sbg}$ with the output from refiner network (projected on to source background).

Our refiner network is trained with two losses: perceptual $\mathcal{L}_{pr}$ and style $\mathcal{L}_{sr}$. $\mathcal{L}_{pr}$ is computed between refined image $r_i$ and the ground-truth target frame $t_i$:
\begin{equation}
  \mathcal{L}_{pr} = \sum_{j}||\phi^j(r_i) - \phi^j(t_i)||_1.
  \label{eq:perceptual_refiner}
\end{equation}

The style loss $\mathcal{L}_{sr}$ measures the statistical error between Gram matrix $G_j^\phi(.)$ constructed from activation maps $\phi_j(.)$ of VGG-19 network, calculated between $r_i$ and $t_i$:
\begin{equation}
  \mathcal{L}_{sr} = \sum_{j}||G_j^\phi(r_i) - G_j^\phi(t_i)||_1.
  \label{eq:style_refiner}
\end{equation}

The total loss is a weighted sum of all the above losses,
\begin{equation}
 \begin{split}
  \mathcal{L}_{total} = \lambda_{pw}\mathcal{L}_{pw} + \lambda_{cw}\mathcal{L}_{cw} + \lambda_{3dw}\mathcal{L}_{3dw} + \lambda_{3df}\mathcal{L}_{3df} + \\
  \lambda_{inp}\mathcal{L}_{inp} + \lambda_{pr}\mathcal{L}_{pr} + \lambda_{sr}\mathcal{L}_{sr}.
  \label{eq:total_loss}
 \end{split}
\end{equation}

In our experiments, we set $\lambda_{pw}=2.5$, $\lambda_{cw}=2.5$, $\lambda_{3dw}=100$, $\lambda_{3df}=100$, $\lambda_{inp}=2.5$, $\lambda_{pr}=4$ and $\lambda_{sr}=1000$.

\section{Evaluation}
\label{sec:evaluation}

\subsection{Implementation Details}
\label{subsec:implementation_details}

\textbf{Dataset.} Our architecture is trained using VoxCeleb dataset \cite{voxceleb} which contains talking-head video sequences extracted from publicly available YouTube videos. Following the pre-processing pipeline in \cite{fomm, pirenderer}, we crop the face regions from original videos with spatial resolution $(256 \times 256)$. We ensure that cropped frames contain natural head movements where the person's face can move freely across all dimensions, however, with a fixed bounding box. We use a total of 16,900 video sequences for training and 477 videos for evaluation, where the each sequence length varies from 64 to 1024 frames. 

\textbf{Evaluation Metrics.} To quantitatively evaluate the realism of output images, we use Fréchet Inception Distance (FID) \cite{fid}, computed between generated image and target image. We additionally use Structural Similarity Index (SSIM) \cite{ssim} and cosine similarity (CSIM) of identity embeddings extracted using ArcFace \cite{arcface}, to evaluate reconstruction quality and identity preservation capability, respectively. Peak signal-to-noise-ratio (PSNR) is also used to measure leakage of noise into output images. Further, for evaluating the accuracy in facial expression (or motion) transfer, we calculate Average Expression Distance (AED) and Average Pose Distance (APD) similar to \cite{pirenderer, styleheat}, and Average Keypoint Distance (AKD) with the method outlined in \cite{fomm}. 

\textbf{Training Details.} As outlined in \cref{subsec:training_strategies}, we pre-train 3D warping and inpainting blocks for a total of 100 epochs, followed by end-to-end training for another 100 epochs. We use ADAM optimizer \cite{adam} with an initial learning rate of $1e^{-4}$. We schedule the learning rate to decay by a factor of 5 at $50^{th}$ epoch.



\subsection{Comparison with state-of-the-art methods}
\label{subsec:comparison}

We evaluate the performance of face re-enactment across two tasks, self-ID re-enactment and cross-ID re-enactment. For self-ID re-enactment, the identity of the individual in the source image matches that of the person in the target video. For cross-ID re-enactment, the identities are different. The evaluation dataset is generated using 477 test videos in VoxCeleb dataset \cite{voxceleb}. For self-ID re-enactment, the first frame of a video sequence is considered as the source and the remaining frames are the target frames. For cross-ID re-enactment, we randomly select two videos from the test set and select one frame for the source and the other video sequence as the target video. For a fair evaluation, we ensure that source-target pairs are kept the same while evaluating each method. We choose the following state-of-the-art one-shot face re-enactment methods for comparison: X2Face \cite{x2face}, Bi-Layer \cite{bilayer}, FOMM \cite{fomm}, PIRenderer \cite{pirenderer}, DaGAN \cite{dagan} and HyperReenact \cite{hyperreenact}. 

\textbf{Quantitative evaluation.} \cref{tab:quantitative_evaluation} shows the quantitative results of our model compared to other methods. In the self-ID re-enactment case, we compute FID, PSNR, SSIM, CSIM and AKD between the generated image and the corresponding ground-truth target image. Even though we do not have a true ground-truth for cross-ID re-enactment, we still use the target frame to compute FID and AKD. Since AED and APD do not rely on individual's identity, we use output-target frame pairs for their computation in both the tasks. Our model achieves the best FID in both the tasks indicating that images generated using our framework are more realistic-looking compared to other methods. Our highest PSNR value means that output contains the least noise leaked into the generated image. Similarly,  our model also achieves the best SSIM and CSIM which translates to better reconstruction quality and preservation of original source identity. The accuracy of expression and motion transfer of our model is also the best as validated by lowest AED, AKD and comparable APD values.

\begin{table}[ht]
  \caption{Quantitative comparisons with state-of-the-art methods.}
  \vspace{-5pt}
  \label{tab:quantitative_evaluation}
  \centering
  \resizebox{\textwidth}{!}{
  \begin{tabular}{@{}c|ccccccc|ccccc@{}}
    \toprule
     & \multicolumn{7}{c|}{Self-ID Re-enactment} & \multicolumn{4}{c}{Cross-ID Re-enactment} \\
     \cmidrule(l){2-12}
     & FID$\downarrow$ & PSNR$\uparrow$ & SSIM$\uparrow$ & CSIM$\uparrow$ & AED$\downarrow$ & APD$\downarrow$ & AKD$\downarrow$ & FID$\downarrow$ & AED$\downarrow$ & APD$\downarrow$ & AKD$\downarrow$ \\
    \midrule
    X2Face \cite{x2face} & 114.27 & 13.9294 & 0.4510 & 0.3710 & 0.2091 & 0.1083 & 7.4112 & 244.46 & 0.3076 & 0.1311 & 9.7703 \\
    Bi-Layer \cite{bilayer} & 186.10 & 6.5490 & 0.3190 & 0.5210 & 0.1367 & 0.0181 & 13.6438 & 254.44 & 0.2385 & 0.0273 & 13.7811\\
    FOMM \cite{fomm} & 58.96 &  21.7255 & 0.7131 & 0.6414 & 0.1172 & 0.0213 & 1.8078 & 236.57 & 0.3059 & 0.2195 & 22.7711 \\
    PIRenderer \cite{pirenderer} & \underline{56.12} & 21.7572 & 0.7061 & 0.6405 & 0.1218 & 0.0294 & 2.4004 & \underline{228.18} & 0.2435 & 0.0427 & \underline{4.6326}\\
    DaGAN \cite{dagan} & 57.04 & \underline{22.0195} & \underline{0.7225} & \underline{0.6420} & \underline{0.1128} & 0.0209 & \underline{1.7426} & 244.45 & 0.3109 & 0.2207 & 23.3367 \\
    HyperReenact \cite{hyperreenact} & 172.34 & 19.2054 &  0.6458 & 0.6045 & 0.1305 & \underline{0.0180} & 13.2605 & 252.14 & \underline{0.2201} & \textbf{0.0234} & 13.2871 \\
    Ours & \textbf{51.87} & \textbf{22.2859} & \textbf{0.7257} & \textbf{0.6615} & \textbf{0.0953} & \textbf{0.0138} & \textbf{1.5193} & \textbf{224.57} & \textbf{0.2174} & \underline{0.0240} & \textbf{3.9293} \\
  \bottomrule
  \end{tabular}}
\end{table}

\begin{figure}[ht]
  \centering
  \begin{subfigure}{\textwidth}
    \centering
    \includegraphics[width=0.85\linewidth]{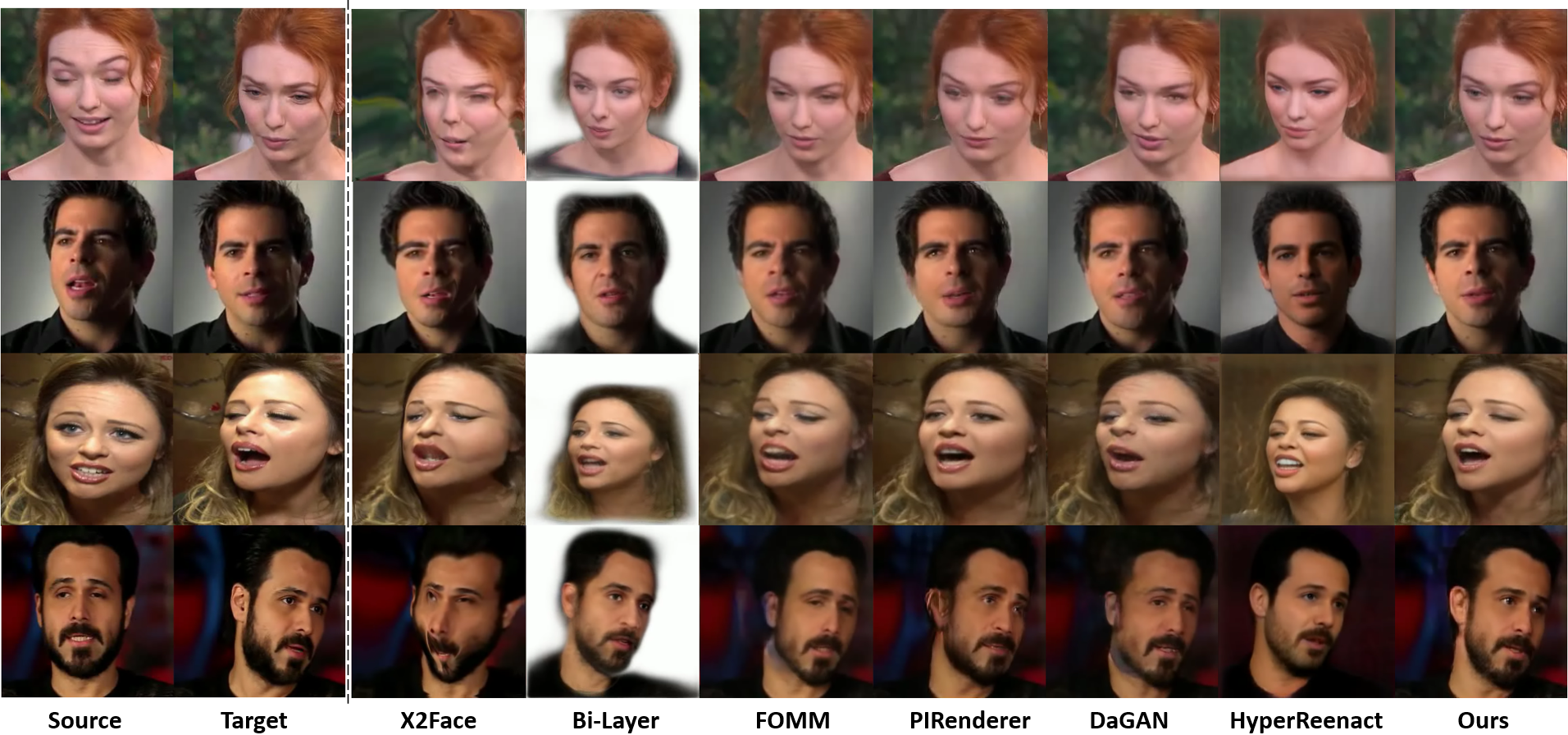}
    \caption{Self-ID re-enactment}
    \label{fig:fig9_a_qualitative_self}
  \end{subfigure}
  \begin{subfigure}{\textwidth}
    \centering
    \includegraphics[width=0.85\linewidth]{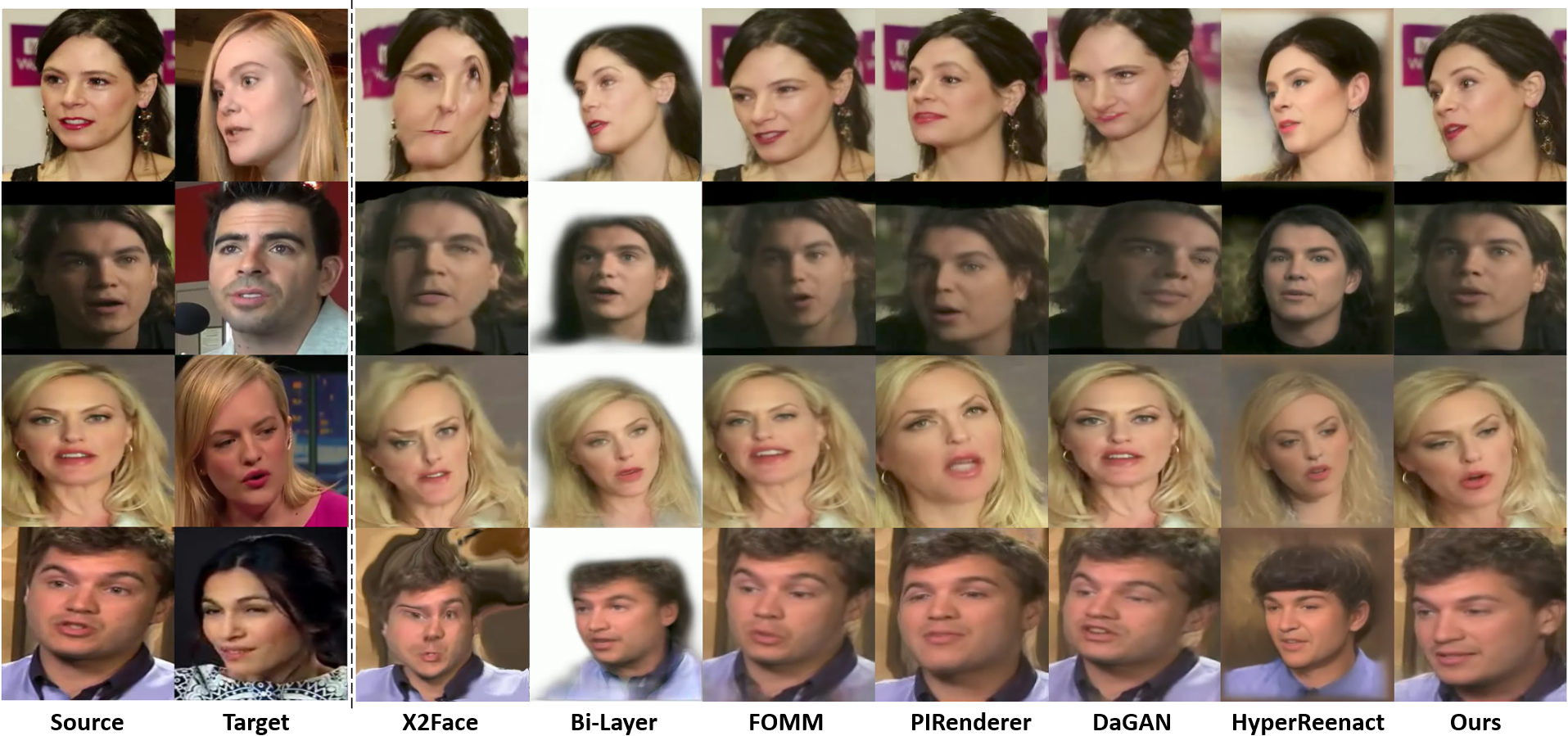}
    \caption{Cross-ID re-enactment}
    \label{fig:fig9_b_qualitative_cross}
  \end{subfigure}
  \vspace{-10pt}
  \vspace{-5pt}
  \caption{Qualitative comparisons with state-of-the-art methods.}
  \label{fig:fig9_qualitative}
\end{figure}

\textbf{Qualitative evaluation.} Next, we compare the quality of generated images through visual analysis. The results of self-ID re-enactment and cross-ID re-enactment are shown in \cref{fig:fig9_a_qualitative_self} and \cref{fig:fig9_b_qualitative_cross}, respectively. As illustrated, our method achieves superior image quality over other methods, specifically in cases where target poses and expressions are significantly different from the source image. The images generated from our method also possess least artifacts and render appropriate background details even when source foreground translates due to motion. X2Face, Bi-Layer, and FOMM face difficulties in rendering artifact free images due to their limitation of 2D warping fields. PIRenderer can produce images with matching expressions, but is not able to transfer appropriate facial poses. DaGAN and HyperReenact work well for misaligned faces, however, the identity of individual in the output image varies significantly from that of source image, specially in cross-ID re-enactment. Moreover, they are unable to render the finer facial details like the pupils, teeth and eye regions. Meanwhile, our method is able to properly render the facial details while maintaining the source identity. Therefore, qualitative evaluation demonstrates that our model outperforms the existing state-of-the-art methods for the task of face re-enactment.

\begin{table}[ht]
  \caption{Ablation Study.}
  \vspace{-15pt}
  \label{tab:ablation_study}
  \centering
  \begin{subtable}{0.32\textwidth}
    \centering
    \label{tab:ablation_study_a}
    \caption{2D flow vs 3D flow field.}
    \vspace{-5pt}
      \begin{tabular}{@{}c|c|c@{}}
        \toprule
        & \multicolumn{2}{c}{FID$\downarrow$} \\
         \cmidrule(l){2-3}
         & Self-ID & Cross-ID\\
        \midrule
        2D flow & 55.27 & 234.91 \\
        3D flow & \textbf{51.87} & \textbf{224.57}\\
      \bottomrule
      \end{tabular}
  \end{subtable}
  \begin{subtable}{0.32\textwidth}
    \centering
    \label{tab:ablation_study_b}
    \caption{Effect of refiner network.}
    \vspace{-5pt}
      \begin{tabular}{@{}c|c|c@{}}
        \toprule
        & \multicolumn{2}{c}{FID$\downarrow$} \\
         \cmidrule(l){2-3}
         & Self-ID & Cross-ID\\
        \midrule
        w/o ref. & 54.42 & 226.61 \\
        w/ ref. & \textbf{51.87} & \textbf{224.57}\\
      \bottomrule
      \end{tabular}
  \end{subtable}
  \begin{subtable}{0.32\textwidth}
    \centering
    \label{tab:ablation_study_c}
    \caption{Effect of Cyclic warp loss.}
    \vspace{-5pt}
      \begin{tabular}{@{}c|c|c@{}}
        \toprule
        & \multicolumn{2}{c}{FID$\downarrow$} \\
         \cmidrule(l){2-3}
         & Self-ID & Cross-ID\\
        \midrule
        w/o $\mathcal{L}_{cw}$ & 53.03 & 225.61 \\
        w/ $\mathcal{L}_{cw}$ & \textbf{51.87} & \textbf{224.57}\\
      \bottomrule
      \end{tabular}
  \end{subtable}
\end{table}

\begin{figure}[ht]
  \centering
  \begin{subfigure}{0.49\linewidth}
    \centering
    \includegraphics[width=0.90\linewidth]{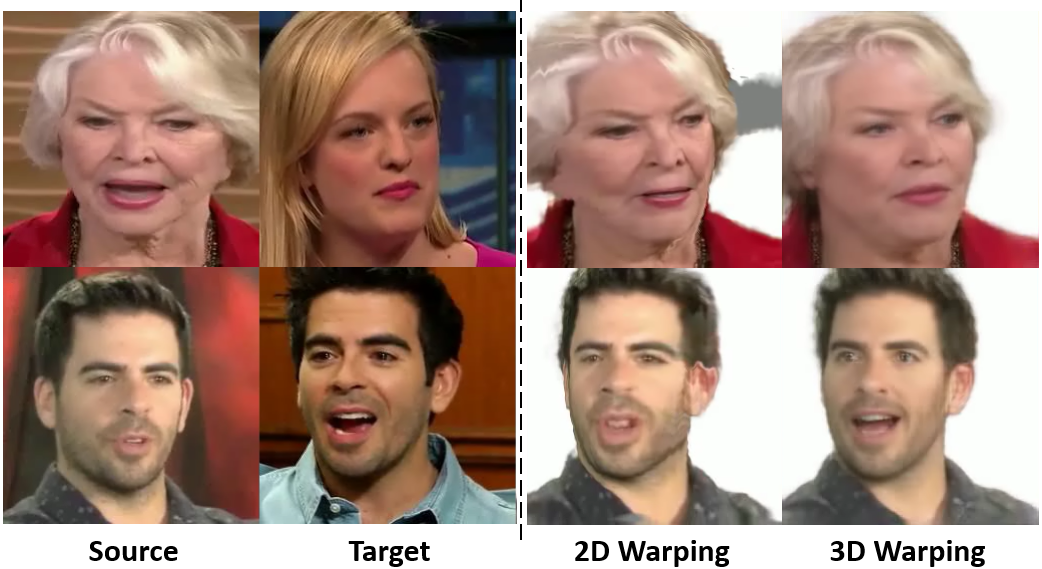}
    \caption{2D vs. 3D warping.}
    \label{fig:fig11_ablation_a}
  \end{subfigure}
    \hfill
  \begin{subfigure}{0.49\linewidth}
    \centering
    \includegraphics[width=0.90\linewidth]{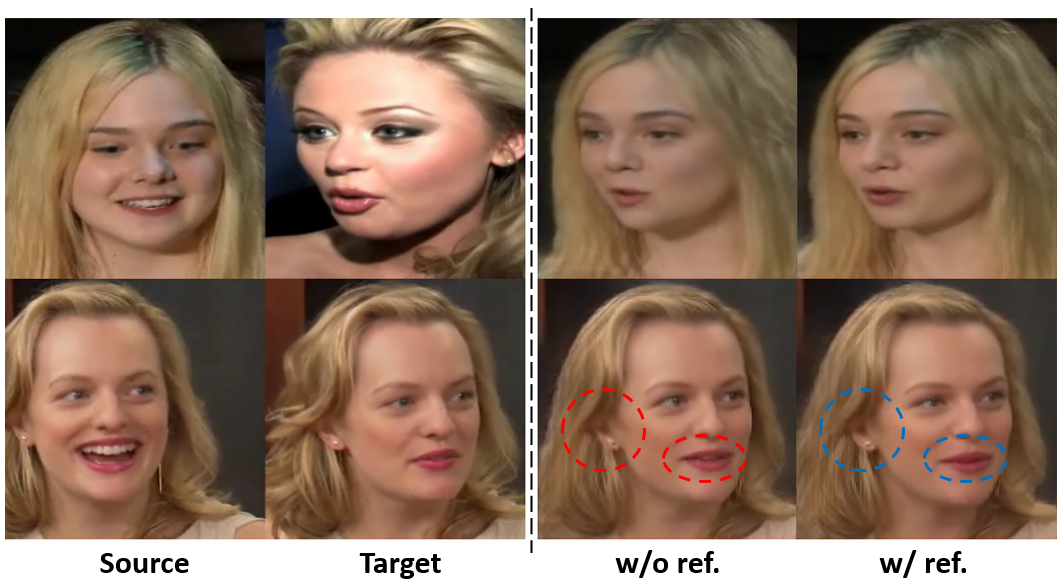}
    \caption{Effect of refiner network.}
    \label{fig:fig11_ablation_b}
  \end{subfigure}
  \begin{subfigure}{0.49\linewidth}
    \centering
    \includegraphics[width=0.90\linewidth]{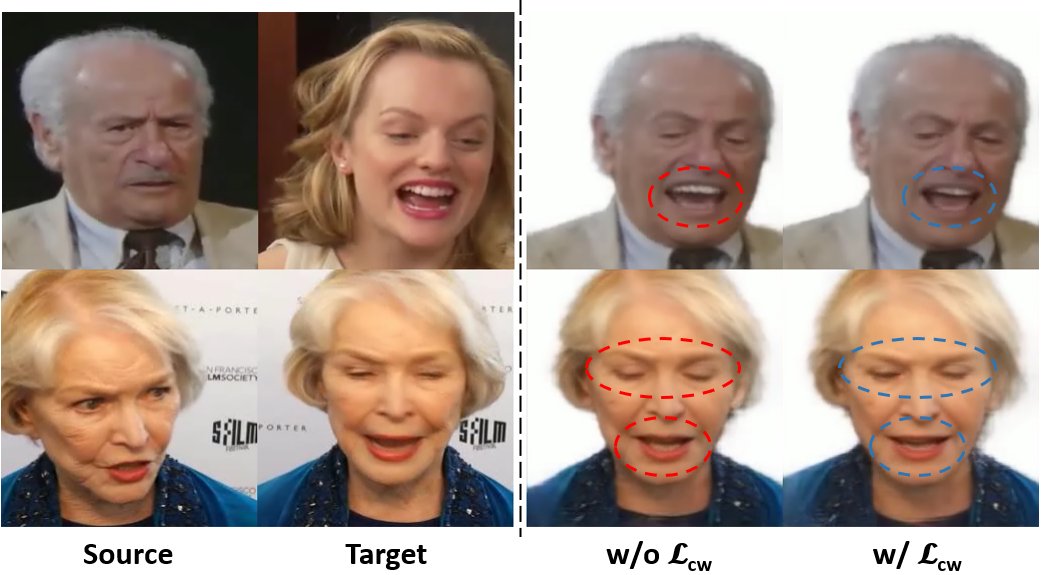}
    \caption{Effect of Cyclic warp loss.}
    \label{fig:fig11_ablation_c}
  \end{subfigure}
  \vspace{-5pt}
  \caption{Visual comparisons for ablation study.}
  \label{fig:fig11_ablation}
\end{figure}

\subsection{Ablation Study}
\label{subsec:ablation_study}

In this section, we evaluate the importance of some key design choices and their effects on the rendering performance of our model. 

\textbf{2D vs. 3D warping.} To evaluate the efficacy of warping in 3D space, we also performed 2D facial flow field estimation to warp the input source image in 2D space. As seen in \cref{tab:ablation_study_a}, images generated using 3D flow achieve lower FID in both cases of self-ID and cross-ID re-enactment indicating that warping in 2D space leads to degradation in image quality. Further, visual analysis in \cref{fig:fig11_ablation_a} reveals that spatially transforming source image using 2D flow field leads to multiple artifacts being generated around the facial region. On the other hand, 3D warping produces a realistic looking image.

\textbf{Refiner network.} We showcase how refiner network (TransUNet) enhances the rendering quality by generating output images directly from the warped images. \cref{fig:fig11_ablation_b} illustrates that there are some visible artifacts in the finer facial regions like lips, cheeks, ears and pupils if we skip the refinement stage. This hypothesis is further validated by \cref{tab:ablation_study_b}, where we see a degradation in overall FID without performing refinement.  

\textbf{Cyclic warp loss.} We compare the rendering performance of 3D warping network by training it with and without the proposed Cyclic warp loss. As shown in \cref{fig:fig11_ablation_c}, we observe that dropping Cyclic warp loss led to the warped image having several facial features (highlighted regions) altered in comparison to the source image. FID values in \cref{tab:ablation_study_c} further showcase the positive impact of regularizing the motion estimation capability using Cyclic warp loss.

\vspace{-10pt}
\section{Conclusion}
We propose 3DFlowRenderer, a novel framework to perform one-shot face re-enactment with robustness to extreme head pose variations by blending the advantages of 2D and 3D methods. Our model estimates dense 3D facial flow fields from input 2D images, warps 3D feature volume of the source image based on motion of the target images, and generates a 2D warped image containing identity of source image with target expressions. Further, we regularize the motion estimation capability of our warping network with proposed Cyclic warp loss, enhancing the accuracy of expression transfer. Besides, our model achieves realistic rendering of finer facial regions while preventing loss and unwanted motion of background pixels. We conduct comprehensive evaluation to demonstrate the re-enactment performance of our network, specifically in challenging scenarios of extreme facial poses and expressions.

\textbf{Ethics consideration.} Face re-enactment technology can generate realistic-looking videos (DeepFakes) of individuals imitating the real ones. Therefore, we acknowledge that such methods can be used for malpractices, like creating fake videos of a specific person without their consent. We strictly condemn such behavior and support spreading social awareness through advancements in research related to face re-enactment. This also aids the development of algorithms targeted at detecting DeepFakes \cite{deepfake_detection1, deepfake_detection2}. It is strictly forbidden to generate and distribute unethical videos using our method.

\clearpage

\section{Supplementary Materials}
\label{supplementary_materials}

In the supplementary materials, we provide the following contents:
\begin{itemize}
\setlength{\parskip}{0cm}
\setlength{\itemsep}{0cm}
\item Detailed network structures of the individual blocks in 3DFlowRenderer.
\item Additional ablation studies to showcase the effectiveness of design choices related to foreground-background separation and windowing during target motion vector generation stage. 
\item Additional qualitative results for the task of self-ID re-enactment and cross-ID re-enactment, evaluated on VoxCeleb dataset \cite{voxceleb}.
\item Visual results showcasing the performance of the proposed TransUNet block in image inpainting task.
\end{itemize}

\subsection{Network structure}
\label{sec:network_structure}

In this section, we share the detailed network structures of the individual blocks in 3DFlowRenderer. 

\begin{figure}[ht]
  \centering
  \includegraphics[height=2.5cm,keepaspectratio]{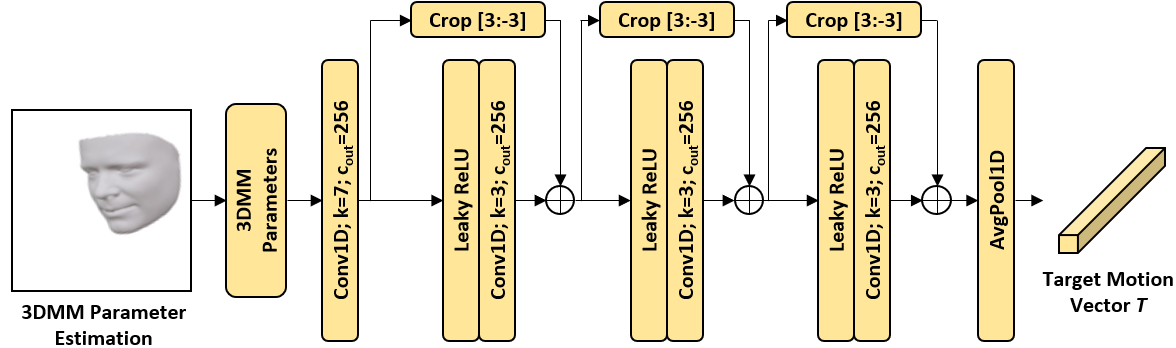}
  \caption{
  The mapping network design.
  }
  \label{fig:fig3_mappingnetwork}
\end{figure}

\textbf{Mapping network.} The network takes a window of 3DMM parameters as an input and applies a series of 1D convolutions and center crop operations. As shown in \cref{fig:fig3_mappingnetwork}, we apply average pooling after multiple stages of affine transformations to generate the target motion vector $T$, similar to PIRenderer \cite{pirenderer}.

\begin{figure}[ht]
  \centering
  
  \begin{subfigure}{0.3\textwidth}
    \centering
    \includegraphics[width=0.8\textwidth]{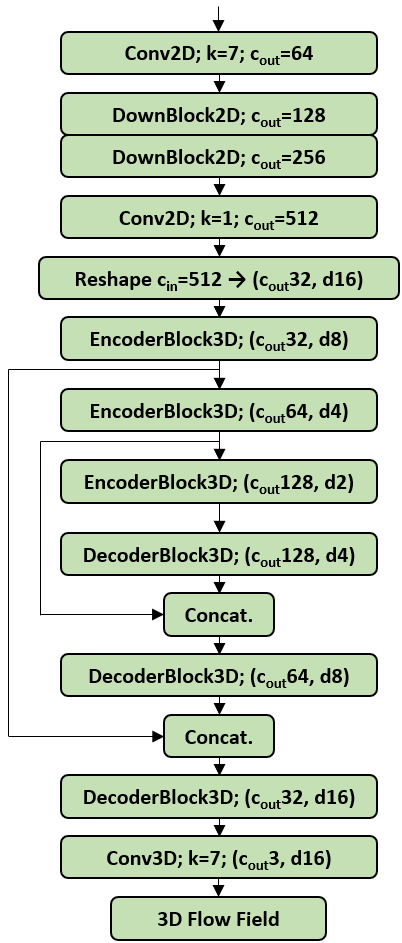}
    \caption{3D Warping Network}
    \label{fig:fig4_a_3dwarpingnetwork}
  \end{subfigure}
    \hfill
  \begin{subfigure}{0.3\textwidth}
    \centering
    \includegraphics[width=0.7\columnwidth]{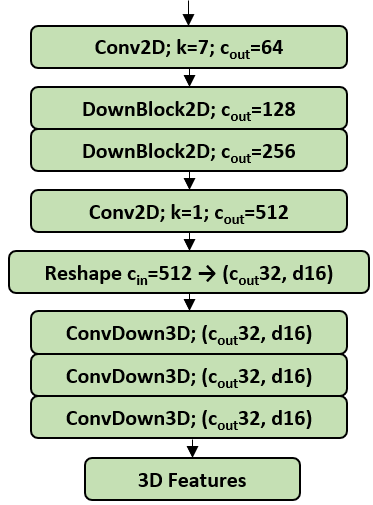}
    \caption{3D Feature Encoder}
    \label{fig:fig4_b_3dencoder}
    \vspace{0.04\linewidth}
    \includegraphics[width=0.7\columnwidth]{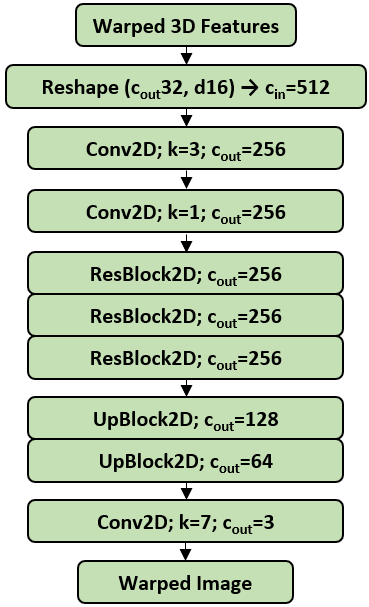}
    \caption{3D Feature Decoder}
    \label{fig:fig4_c_3ddecoder}
  \end{subfigure}
    \hfill 
  \begin{subfigure}{0.3\textwidth}
    \centering
    \includegraphics[width=0.8\textwidth]{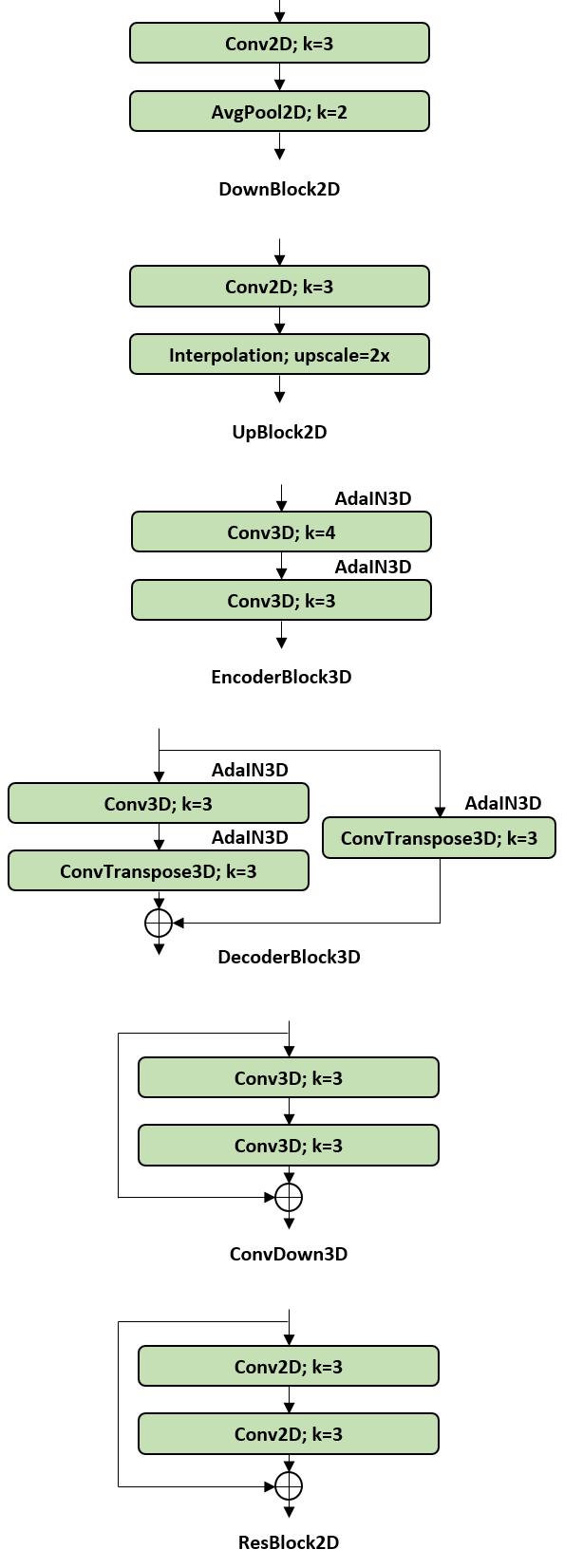}
    \caption{Internal Blocks}
    \label{fig:fig4_d_internalblocks}
  \end{subfigure}

  \caption{Architecture of individual components in the 3D warping stage.}
  \label{fig:fig4_3dwarping}
\end{figure}

\textbf{3D warping network.} \cref{fig:fig4_a_3dwarpingnetwork} illustrates the overall design of our 3D warping network. It follows an auto-encoder design, where we first downsample the input source foreground and reshape the color information into depth and channel space. Then, we apply a series of 3D encoder blocks which downsamples the spatial and depth dimensions after each stage. Here, we inject the target motion into source image statistics by employing AdaIN operation (in 3D) after every convolution block inside the 3D encoder, as illustrated in \cref{fig:fig4_d_internalblocks}. To account for the additional dimension of depth, we modify the AdaIN by using 3D scale and bias parameters instead of their 2D counterparts. This is followed by applying a series of 3D decoder blocks performing transposed convolution on the previously obtained encoded feature maps, concatenated with the output of the previous block. We inject the target motion again after every convolution block using AdaIN. The output of this network is the dense 3D facial flow field.

\textbf{3D feature encoder.} To generate a set of 3D appearance features from 2D source image, we use the network illustrated in \cref{fig:fig4_b_3dencoder}. Similar to 3D warping network, it first reshapes the input into depth and channel space after downsampling. Then, we apply 3 encoder blocks titled "ConvDown3D" (\cref{fig:fig4_d_internalblocks}), which contains a pair of 3D convolutions in series along with an additive skip connection. At each convolution layer, we keep the depth dimension constant.

\textbf{3D feature decoder.} The input to 3D feature decoder is a 3D feature volume. It first projects the 3D features back into 2D and applies a series of 2D residual blocks (\cref{fig:fig4_d_internalblocks}) to obtain low-resolution 2D feature maps. Here, the channel dimension after each layer is kept constant. Finally, we upsample the feature maps to match the spatial dimension of the 2D source image and generate the RGB warped image.

\begin{figure}[ht]
  \centering
  \includegraphics[width=\textwidth,keepaspectratio]{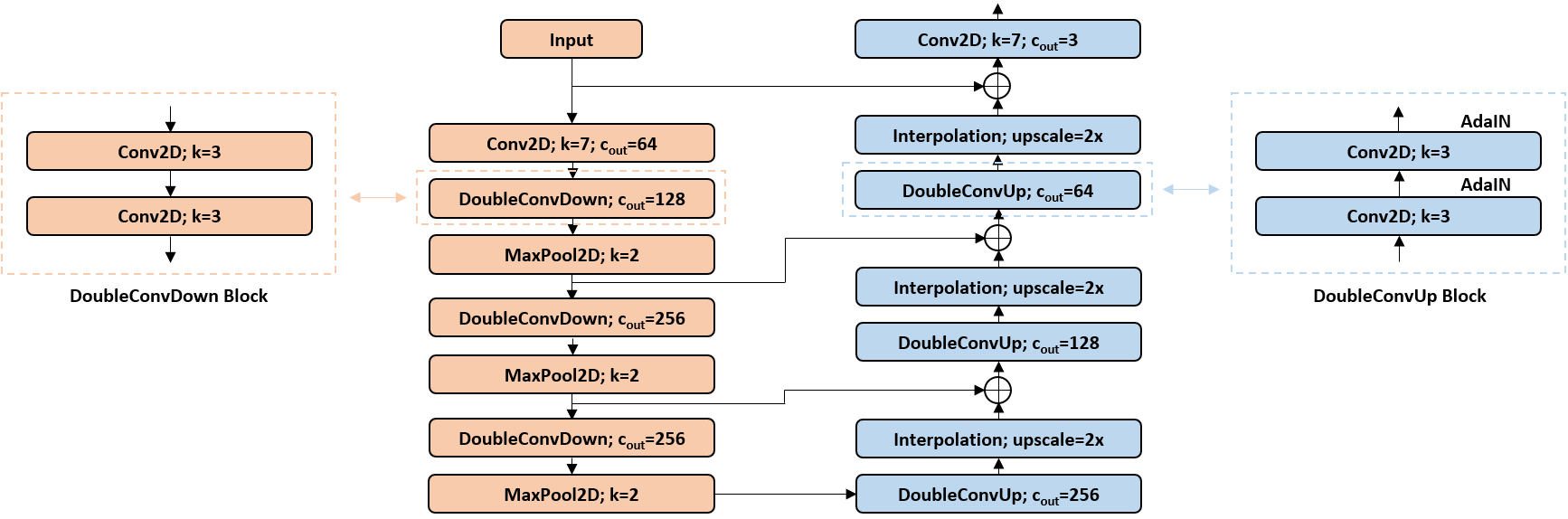}
  \caption{
  The TransUNet block design.
  }
  \label{fig:fig5_transunet}
\end{figure}

\textbf{TransUNet block.} The design of our proposed TransUNet block is illustrated in \cref{fig:fig5_transunet}. It is based on UNet \cite{unet} with several modifications. The encoder part applies a series of 2D convolutions followed by max pooling, downsampling the spatial resolution by a factor of 2 after each layer. The decoder part again applies a series of 2D convolutions and performs nearest neighbor interpolation after each layer. This maintains the overall spatial resolution of the input. To ensure realistic rendering of finer facial details like pupils, eyes, teeth, lips, etc., we inject the target motion vector into the warped (and source) facial statistics by employing AdaIN operation after every convolution block in the decoder. This key modification enables the refinement of the warped facial expressions and removes any unwanted artifacts caused by the warping operation. The skip connections also help in preserving the source facial textures without tampering the identity of the individual in the source image. Due to its universal design, we use the same network with different weights for performing the task of image inpainting.

\subsection{Additional ablation studies}
\label{sec:additional_ablation_studies}

In this section, we evaluate the importance of two additional design choices and their effects on the rendering performance of our model. The evaluation dataset is generated using 477 test videos in VoxCeleb dataset \cite{voxceleb} across two tasks, self-ID re-enactment and cross-ID re-enactment. For the task of self-ID re-enactment, the first frame of a test video is considered as the source and the remaining frames are the target frames. We randomly selected one frame from a video as source and another video sequence as target video for cross-ID re-enactment. To ensure fairness in evaluating various versions of our model during cross-ID re-enactment, we select the same source-target pairs.

    
\begin{table}[ht]
  \caption{Ablation Study.}
  \vspace{-15pt}
  \label{tab:ablation_study_supp}
  \centering
  \begin{subtable}{0.49\textwidth}
    \centering
    \label{tab:ablation_study_a_supp}
    \caption{Effect of rendering without foreground-background separation.}
    \vspace{-5pt}
      \begin{tabular}{@{}c|cc|c@{}}
        \toprule
        & \multicolumn{2}{c|}{Self-ID} & \multicolumn{1}{c}{Cross-ID} \\
         \cmidrule(l){2-4}
         & FID$\downarrow$ & PSNR$\uparrow$ & FID$\downarrow$\\
        \midrule
        w/o sep. & 81.97 & 16.4695 & 241.91 \\
        w/ sep. & \textbf{51.87} & \textbf{22.2859} & \textbf{224.57}\\
      \bottomrule
      \end{tabular}
  \end{subtable}
    \hfill
  \begin{subtable}{0.49\textwidth}
    \centering
    \label{tab:ablation_study_b_supp}
    \caption{Effect of window size $k$ for target motion vector generation.}
    \vspace{-5pt}
      \begin{tabular}{@{}c|c|c@{}}
        \toprule
        & \multicolumn{2}{c}{FID $\downarrow$} \\
         \cmidrule(l){2-3}
         & Self-ID & Cross-ID\\
        \midrule
        $k=0$ & 54.27 & 227.24 \\
        $k=13$ & \textbf{51.87} & \textbf{224.57}\\
      \bottomrule
      \end{tabular}
  \end{subtable}
\end{table}

\begin{figure}[t]
  \centering
  \begin{subfigure}{0.7\linewidth}
    \centering
    \includegraphics[width=0.96\linewidth]{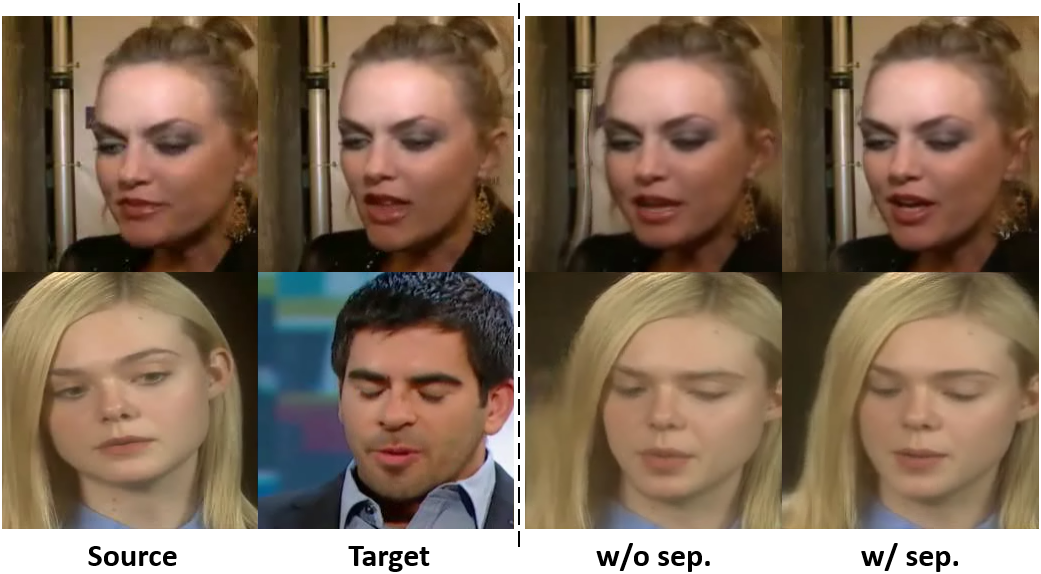}
    \caption{Effect of rendering without foreground-background separation.}
    \label{fig:fig11_ablation_a_additional}
  \end{subfigure}
  \begin{subfigure}{\linewidth}
    \centering
    \includegraphics[width=0.96\linewidth]{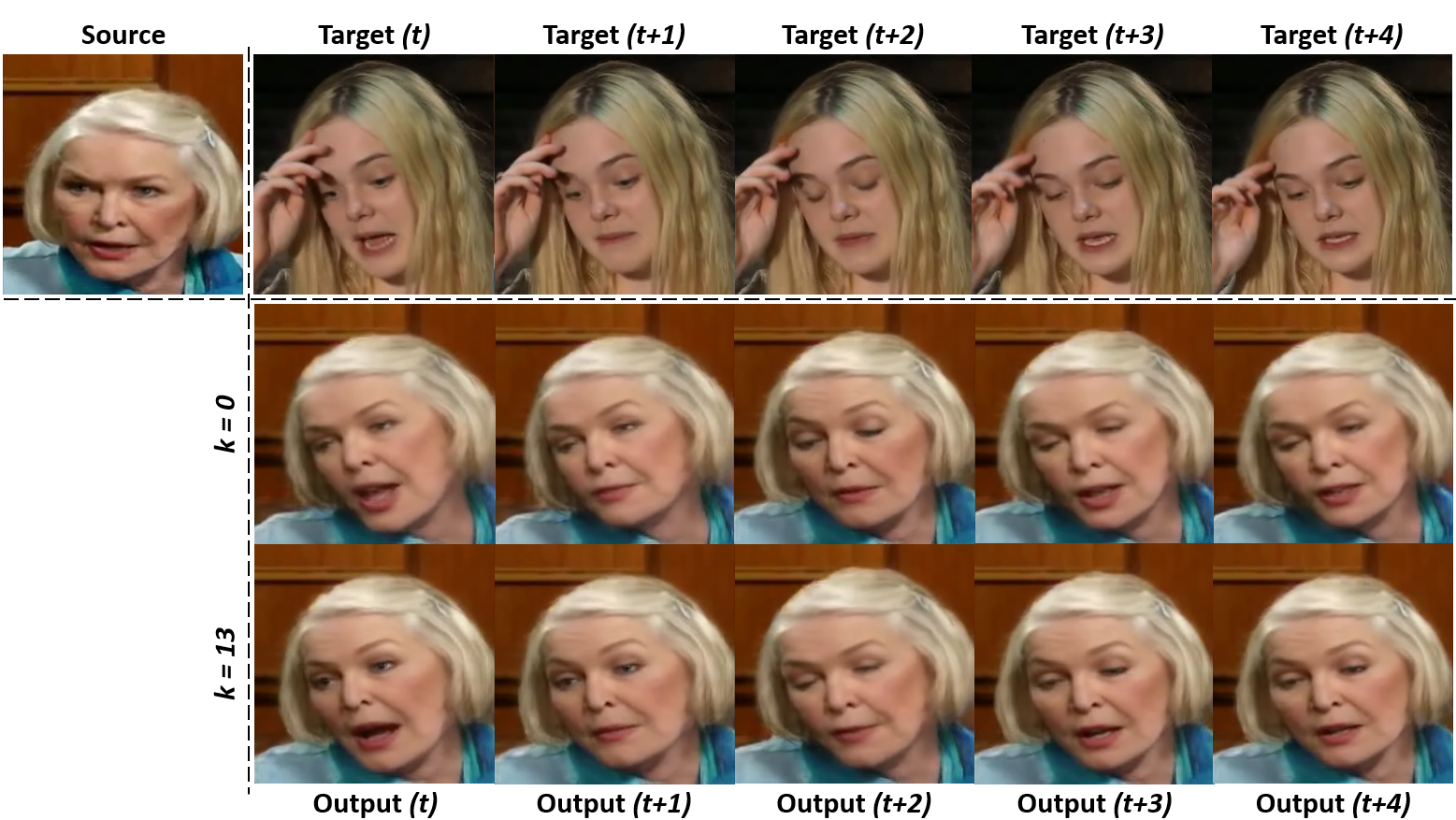}
    \caption{Effect of window size $k$ on rendering quality of 5 consecutive frames.}
    \label{fig:fig11_ablation_b_additional}
  \end{subfigure}
  \vspace{-5pt}
  \caption{Visual comparisons for additional ablation studies.}
  \label{fig:fig11_ablation_additional}
\end{figure} 

\textbf{Rendering without foreground-background separation.} We evaluate the improvement in rendering performance by applying the proposed pipeline, which separately renders the foreground region first, followed by inpainting the background. \cref{tab:ablation_study_a_supp} shows the quantitative results of our model trained without background separation against the proposed model which performs background extraction using RVM \cite{rvm}. We compute Fréchet Inception Distance (FID) \cite{fid} and Peak signal-to-noise-ratio (PSNR) between the generated image and the corresponding ground-truth target image for self-ID re-enactment. For cross-ID re-enactment, we again use output-target frame pairs for computation since we do not have the true ground-truths. For both the tasks, the model trained without the separation achieved poor FID value, indicating that the realism of output is reduced by rendering the entire scene instead of just the facial foreground region. Similarly, a low PSNR value is directly correlated to the leakage of noise into the facial foreground region of the output image from the neighboring background pixels. Visual comparison illustrated in \cref{fig:fig11_ablation_a_additional} further validates the hypothesis, as noticeable movement in background pixels can be seen in the output image generated without separation (first row). On the contrary, there is no movement of background artifacts if we render the foreground and background separately. In the case of occlusion (\cref{fig:fig11_ablation_a_additional}, second row), where the individual in source image has partially visible hair texture, the image generated without separation contains random noise when the face translates due to the target motion. This limitation is rectified by using our proposed method of rendering the foreground separately.

\textbf{Window size $k$ for target motion vector generation.} Our method relies on Deep3DFaceRecon \cite{deep3dfacerecon} for predicting 3DMM parameters which is susceptible to estimation errors. To average out these accumulated errors and prevent degradation in performance, we pass a window (size $k=13$) of continuous frames to the estimation network. For the purpose of studying the improvement in rendering quality, we ablate the network by training it using only a single target frame $(k=0)$ as the input to 3DMM parameter estimation network. The higher FID values of model trained with $k=0$ in \cref{tab:ablation_study_b_supp} show that the generated output lacks realism due to poor estimation of 3DMM parameters using only a single frame, for both the tasks of self-ID and cross-ID re-enactment. Further, \cref{fig:fig11_ablation_b_additional} visually compares the rendered output for a source image and 5 consecutive target frames for $k=0$ and $k=13$. We notice errors in motion transfer, specifically the motion of blinking, where in target frames $(t+1)$, $(t+2)$ and $(t+4)$ only one eyelid appears to be closed for $k=0$ even though the ground-truth has both eyelids closed. This is not the case for the output generated using model with $k=13$. Therefore, both quantitative comparison and visual analysis reveals the effectiveness of passing a window of continuous frames for target motion vector generation instead of a single frame.

\subsection{Additional qualitative results}
\label{sec:additional_qualitative_results}

\begin{figure}[!b]
  \centering
    \includegraphics[width=0.96\linewidth]{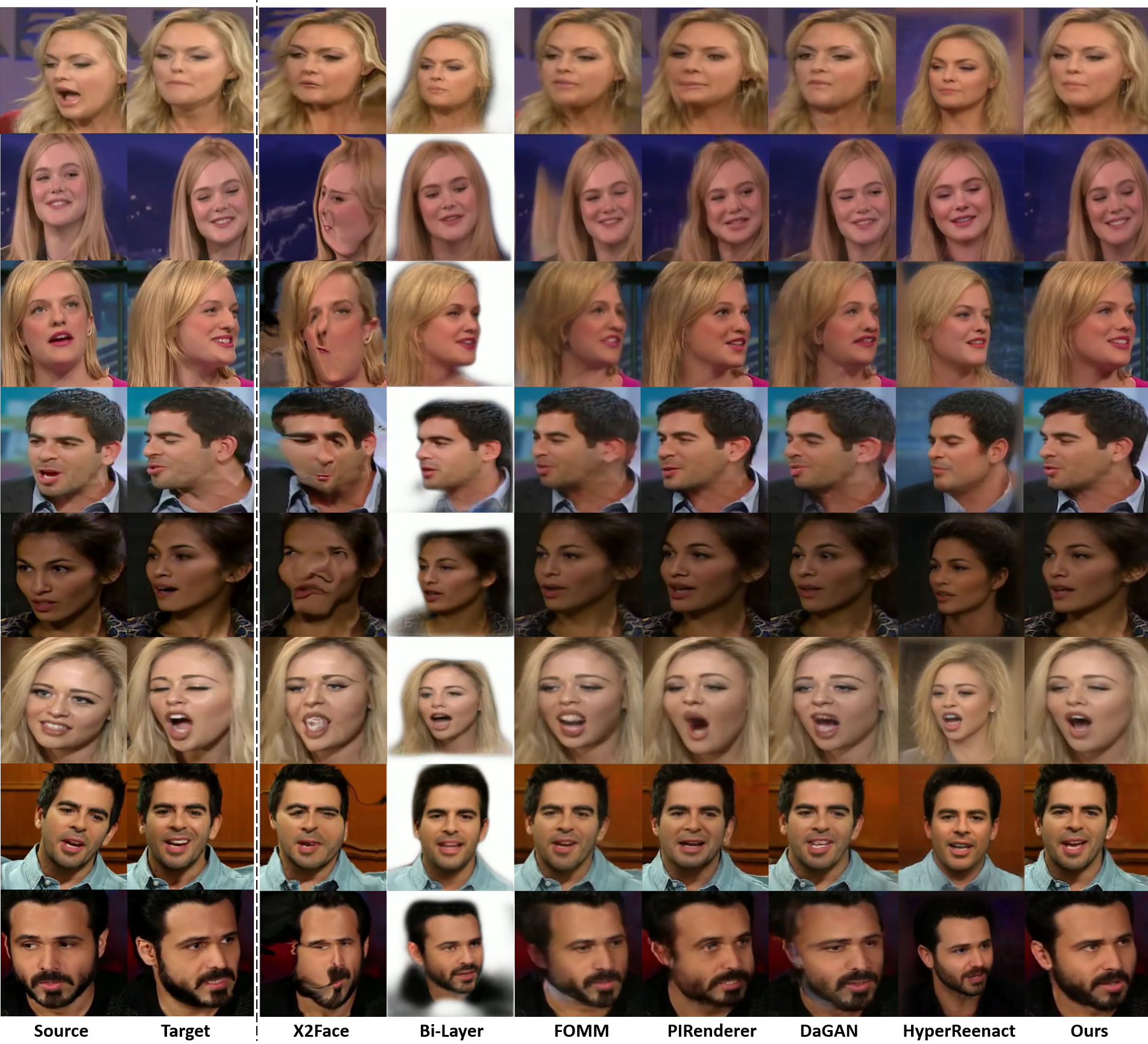}
    \caption{Qualitative comparisons with state-of-the-art methods for the task of self-ID re-enactment.}
    \label{fig:fig9_a_qualitative_self_additional}
\end{figure}

\begin{figure}[t]
  \centering
    \includegraphics[width=0.96\linewidth]{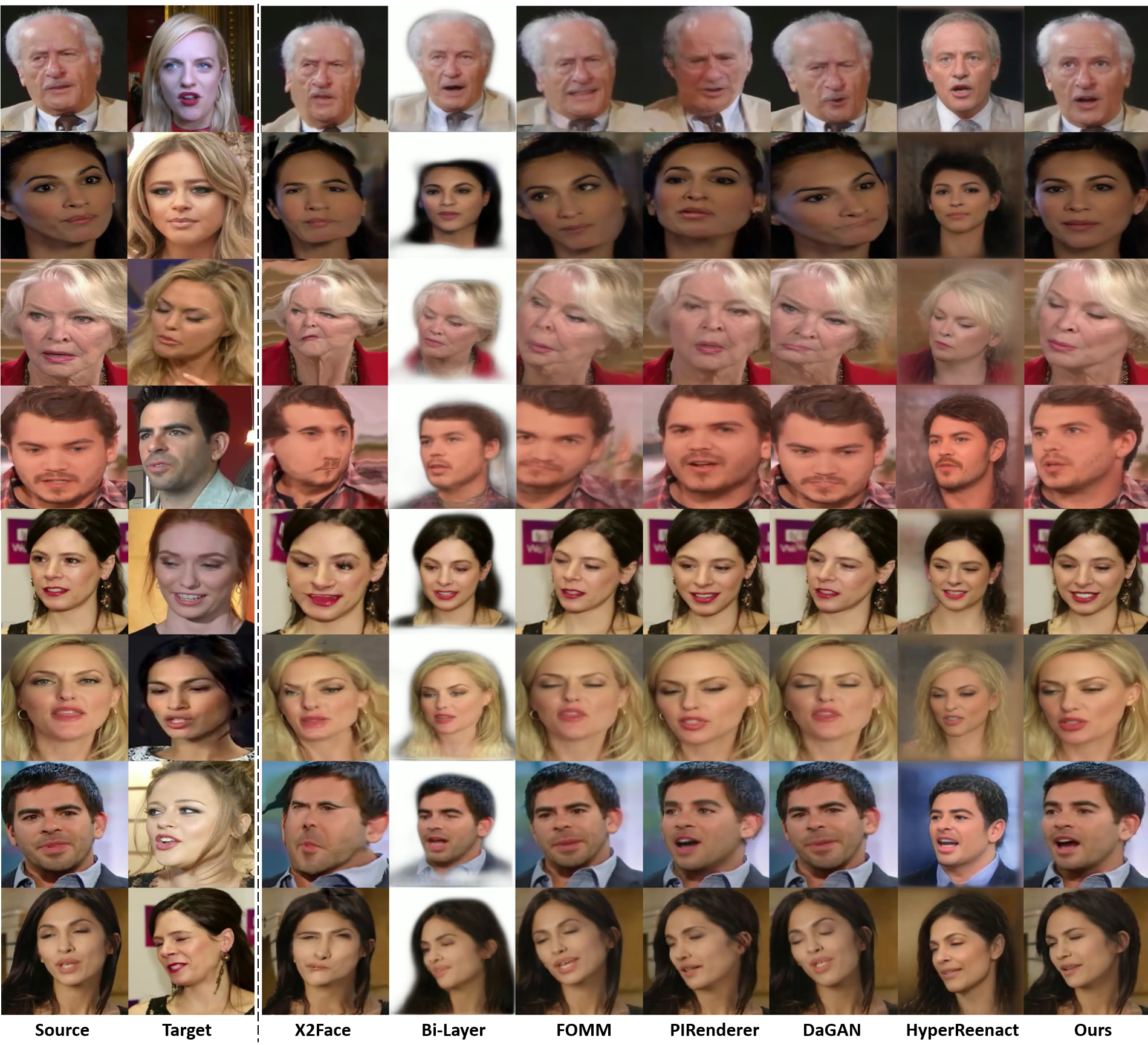}
    \caption{Qualitative comparisons with state-of-the-art methods for the task of cross-ID re-enactment.}
    \label{fig:fig9_b_qualitative_cross_additional}
\end{figure}

We showcase some more qualitative results of our model compared to existing state-of-the-art methods (X2Face \cite{x2face}, Bi-Layer \cite{bilayer}, FOMM \cite{fomm}, PIRenderer \cite{pirenderer}, DaGAN \cite{dagan} and HyperReenact \cite{hyperreenact}) on the task of self-ID re-enactment (\cref{fig:fig9_a_qualitative_self_additional}) and cross-ID re-enactment (\cref{fig:fig9_b_qualitative_cross_additional}). We use the same evaluation dataset outlined in \cref{sec:additional_ablation_studies}. Visual comparisons reveal that the images generated using our method are more realistic-looking when compared to other methods, and contain the least amount of unwanted artifacts. 2D warping methods like X2Face, Bi-layer, FOMM and PIRenderer are limited to perceiving extreme poses. They produce blurry textures and unnatural looking facial details on multiple occasions. DaGAN is able to model facial expressions, but fails to match the target pose in several cross-ID re-enactment cases (second, third, fourth and seventh rows in \cref{fig:fig9_b_qualitative_cross_additional}). HyperReenact renders a realistic looking face even in extreme poses, but fails to preserve the identity of the individual in the source image. It produces a synthetic looking face because of its reliance on StyleGAN2 generator \cite{stylegan2}. 

\subsection{Visual results for image inpainting task}

\begin{figure}[t]
  \centering
  \includegraphics[width=0.5\textwidth]{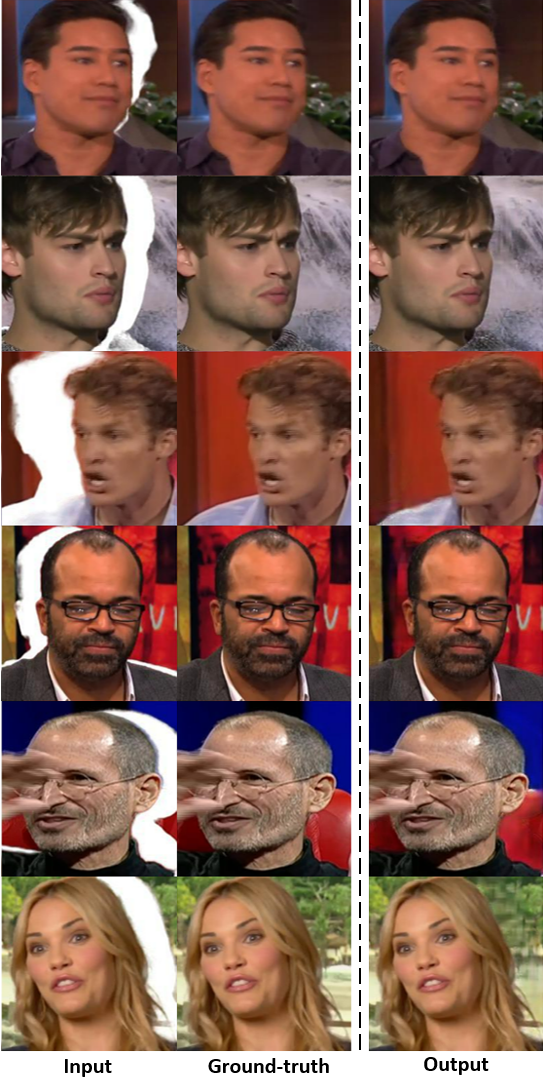}
  \caption{
  Visual results for image inpainting task.
  }
  \label{fig:fig12_inpainting}
\end{figure}

In this section, we showcase the visual results of the image inpainting task performed using the proposed TransUNet block. The images in \cref{fig:fig12_inpainting} are taken from the evaluation dataset outlined in \cref{sec:additional_ablation_studies}. We first sample a source-target image pair containing the same individual, and project the target foreground on to the source background (separated using RVM \cite{rvm}). This generates the input image containing several blank spaces due to the face translation. We then feed these images into the TransUNet block trained on the image inpainting task and compare the output to the corresponding ground-truth. \cref{fig:fig12_inpainting} showcases that TransUNet is able to generate high-frequency details in the background, specifically in the first, second, and the sixth rows. It can also inpaint the background with matching colors even when the source translation is large (third row). Further, it can generate multiple colors and geometric shapes as illustrated in the fourth and fifth rows, adding the realism to the output images.


\clearpage

\clearpage

\clearpage



%
%
\bibliographystyle{splncs04}
\bibliography{main}

\begin{thebibliography}{10}
\providecommand{\url}[1]{\texttt{#1}}
\providecommand{\urlprefix}{URL }
\providecommand{\doi}[1]{https://doi.org/#1}

\bibitem{hyperreenact}
Bounareli, S., Tzelepis, C., Argyriou, V., Patras, I., Tzimiropoulos, G.: Hyperreenact: One-shot reenactment via jointly learning to refine and retarget faces. In: Proceedings of the IEEE/CVF International Conference on Computer Vision (ICCV) (2023)

\bibitem{facewarehouse}
Cao, C., Weng, Y., Zhou, S., Tong, Y., Zhou, K.: Facewarehouse: A 3d facial expression database for visual computing. IEEE Transactions on Visualization and Computer Graphics  \textbf{20}(3),  413--425 (2014). \doi{10.1109/TVCG.2013.249}

\bibitem{deepfake_detection2}
Corvi, R., Cozzolino, D., Zingarini, G., Poggi, G., Nagano, K., Verdoliva, L.: On the detection of synthetic images generated by diffusion models. In: ICASSP 2023 - 2023 IEEE International Conference on Acoustics, Speech and Signal Processing (ICASSP). pp.~1--5 (2023). \doi{10.1109/ICASSP49357.2023.10095167}

\bibitem{deepfake_detection1}
Cozzolino, D., Rössler, A., Thies, J., Nießner, M., Verdoliva, L.: Id-reveal: Identity-aware deepfake video detection. In: 2021 IEEE/CVF International Conference on Computer Vision (ICCV). pp. 15088--15097 (2021). \doi{10.1109/ICCV48922.2021.01483}

\bibitem{arcface}
Deng, J., Guo, J., Yang, J., Xue, N., Kotsia, I., Zafeiriou, S.: Arcface: Additive angular margin loss for deep face recognition. IEEE Transactions on Pattern Analysis and Machine Intelligence  \textbf{44}(10),  5962–5979 (Oct 2022). \doi{10.1109/tpami.2021.3087709}, \url{http://dx.doi.org/10.1109/TPAMI.2021.3087709}

\bibitem{deep3dfacerecon}
Deng, Y., Yang, J., Xu, S., Chen, D., Jia, Y., Tong, X.: Accurate 3d face reconstruction with weakly-supervised learning: From single image to image set. In: IEEE Computer Vision and Pattern Recognition Workshops (2019)

\bibitem{headgan}
Doukas, M.C., Zafeiriou, S., Sharmanska, V.: Headgan: One-shot neural head synthesis and editing. In: Proceedings of the IEEE/CVF International Conference on Computer Vision (ICCV). pp. 14398--14407 (October 2021)

\bibitem{dynamicnerf}
Gafni, G., Thies, J., Zollh{\"o}fer, M., Nie{\ss}ner, M.: Dynamic neural radiance fields for monocular 4d facial avatar reconstruction. In: Proceedings of the IEEE/CVF Conference on Computer Vision and Pattern Recognition (CVPR). pp. 8649--8658 (June 2021)

\bibitem{stylesync}
Guan, J., Zhang, Z., Zhou, H., HU, T., Wang, K., He, D., Feng, H., Liu, J., Ding, E., Liu, Z., Wang, J.: Stylesync: High-fidelity generalized and personalized lip sync in style-based generator. In: Proceedings of the IEEE/CVF Conference on Computer Vision and Pattern Recognition (CVPR) (2023)

\bibitem{fid}
Heusel, M., Ramsauer, H., Unterthiner, T., Nessler, B., Hochreiter, S.: Gans trained by a two time-scale update rule converge to a local nash equilibrium. In: Proceedings of the 31st International Conference on Neural Information Processing Systems. p. 6629–6640. NIPS'17, Curran Associates Inc., Red Hook, NY, USA (2017)

\bibitem{dagan}
Hong, F.T., Zhang, L., Shen, L., Xu, D.: Depth-aware generative adversarial network for talking head video generation. In: Proceedings of the IEEE/CVF Conference on Computer Vision and Pattern Recognition (CVPR). pp. 3397--3406 (June 2022)

\bibitem{headnerf}
Hong, Y., Peng, B., Xiao, H., Liu, L., Zhang, J.: Headnerf: A real-time nerf-based parametric head model. In: {IEEE/CVF} Conference on Computer Vision and Pattern Recognition (CVPR) (2022)

\bibitem{adain}
Huang, X., Belongie, S.: Arbitrary style transfer in real-time with adaptive instance normalization. In: ICCV (2017)

\bibitem{bfm}
IEEE: A 3D Face Model for Pose and Illumination Invariant Face Recognition (2009)

\bibitem{perceptual_loss}
Johnson, J., Alahi, A., Fei-Fei, L.: Perceptual losses for real-time style transfer and super-resolution. In: Leibe, B., Matas, J., Sebe, N., Welling, M. (eds.) Computer Vision -- ECCV 2016. pp. 694--711. Springer International Publishing, Cham (2016)

\bibitem{megafr}
Kang, W., Lee, G., Koo, H.I., Cho, N.I.: One-shot face reenactment on megapixels (2022)

\bibitem{stylegan}
Karras, T., Laine, S., Aila, T.: A style-based generator architecture for generative adversarial networks. In: Proceedings of the IEEE/CVF Conference on Computer Vision and Pattern Recognition (CVPR) (June 2019)

\bibitem{stylegan2}
Karras, T., Laine, S., Aittala, M., Hellsten, J., Lehtinen, J., Aila, T.: Analyzing and improving the image quality of stylegan. In: Proceedings of the IEEE/CVF Conference on Computer Vision and Pattern Recognition (CVPR) (June 2020)

\bibitem{adam}
Kingma, D.P., Ba, J.: Adam: A method for stochastic optimization (2017)

\bibitem{hidenerf}
Li, W., Zhang, L., Wang, D., Zhao, B., Wang, Z., Chen, M., Zhang, B., Wang, Z., Bo, L., Li, X.: One-shot high-fidelity talking-head synthesis with deformable neural radiance field. In: Proceedings of the IEEE/CVF Conference on Computer Vision and Pattern Recognition (CVPR). pp. 17969--17978 (June 2023)

\bibitem{rvm}
Lin, S., Yang, L., Saleemi, I., Sengupta, S.: Robust high-resolution video matting with temporal guidance. In: Proceedings of the IEEE/CVF Winter Conference on Applications of Computer Vision (WACV). pp. 238--247 (January 2022)

\bibitem{otavatar}
Ma, Z., Zhu, X., Qi, G.J., Lei, Z., Zhang, L.: Otavatar: One-shot talking face avatar with controllable tri-plane rendering. In: Proceedings of the IEEE/CVF Conference on Computer Vision and Pattern Recognition (CVPR). pp. 16901--16910 (June 2023)

\bibitem{nerf}
Mildenhall, B., Srinivasan, P.P., Tancik, M., Barron, J.T., Ramamoorthi, R., Ng, R.: Nerf: Representing scenes as neural radiance fields for view synthesis. In: ECCV (2020)

\bibitem{voxceleb}
Nagrani, A., Chung, J.S., Zisserman, A.: {VoxCeleb: A Large-Scale Speaker Identification Dataset}. In: Proc. Interspeech 2017. pp. 2616--2620 (2017). \doi{10.21437/Interspeech.2017-950}

\bibitem{oorloff2023oneshot}
Oorloff, T., Yacoob, Y.: Robust one-shot face video re-enactment using hybrid latent spaces of stylegan2. In: Proceedings of the IEEE/CVF International Conference on Computer Vision (ICCV). pp. 20947--20957 (October 2023)

\bibitem{dpe}
Pang, Y., Zhang, Y., Quan, W., Fan, Y., Cun, X., Shan, Y., Yan, D.M.: Dpe: Disentanglement of pose and expression for general video portrait editing. In: Proceedings of the IEEE/CVF Conference on Computer Vision and Pattern Recognition (CVPR). pp. 427--436 (June 2023)

\bibitem{nerfies}
Park, K., Sinha, U., Barron, J.T., Bouaziz, S., Goldman, D.B., Seitz, S.M., Martin-Brualla, R.: Nerfies: Deformable neural radiance fields. ICCV  (2021)

\bibitem{5279762}
Paysan, P., Knothe, R., Amberg, B., Romdhani, S., Vetter, T.: A 3d face model for pose and illumination invariant face recognition. In: 2009 Sixth IEEE International Conference on Advanced Video and Signal Based Surveillance. pp. 296--301 (2009). \doi{10.1109/AVSS.2009.58}

\bibitem{pirenderer}
Ren, Y., Li, G., Chen, Y., Li, T.H., Liu, S.: Pirenderer: Controllable portrait image generation via semantic neural rendering. In: Proceedings of the IEEE/CVF International Conference on Computer Vision (ICCV). pp. 13759--13768 (October 2021)

\bibitem{unet}
Ronneberger, O., Fischer, P., Brox, T.: U-net: Convolutional networks for biomedical image segmentation (2015)

\bibitem{interfacegan}
Shen, Y., Gu, J., Tang, X., Zhou, B.: Interpreting the latent space of gans for semantic face editing. In: Proceedings of the IEEE/CVF Conference on Computer Vision and Pattern Recognition (CVPR) (June 2020)

\bibitem{animatingarbit}
Siarohin, A., Lathuilière, S., Tulyakov, S., Ricci, E., Sebe, N.: Animating arbitrary objects via deep motion transfer. In: The IEEE Conference on Computer Vision and Pattern Recognition (CVPR) (June 2019)

\bibitem{fomm}
Siarohin, A., Lathuilière, S., Tulyakov, S., Ricci, E., Sebe, N.: First order motion model for image animation. In: Conference on Neural Information Processing Systems (NeurIPS) (December 2019)

\bibitem{vgg19}
Simonyan, K., Zisserman, A.: Very deep convolutional networks for large-scale image recognition. In: International Conference on Learning Representations (2015)

\bibitem{next3d}
Sun, J., Wang, X., Wang, L., Li, X., Zhang, Y., Zhang, H., Liu, Y.: Next3d: Generative neural texture rasterization for 3d-aware head avatars. In: CVPR (2023)

\bibitem{fenerf}
Sun, J., Wang, X., Zhang, Y., Li, X., Zhang, Q., Liu, Y., Wang, J.: Fenerf: Face editing in neural radiance fields. In: Proceedings of the IEEE/CVF Conference on Computer Vision and Pattern Recognition. pp. 7672--7682 (2022)

\bibitem{stylerig}
Tewari, A., Elgharib, M., Bharaj, G., Bernard, F., Seidel, H.P., P{\'e}rez, P., Z{\"o}llhofer, M., Theobalt, C.: Stylerig: Rigging stylegan for 3d control over portrait images, cvpr 2020. In: {IEEE} Conference on Computer Vision and Pattern Recognition (CVPR). {IEEE} (june 2020)

\bibitem{facegan}
Tripathy, S., Kannala, J., Rahtu, E.: Facegan: Facial attribute controllable reenactment gan. In: Proceedings of the IEEE/CVF Winter Conference on Applications of Computer Vision (WACV) (2021)

\bibitem{styleavatar}
Wang, L., Zhao, X., Sun, J., Zhang, Y., Zhang, H., Yu, T., Liu, Y.: Styleavatar: Real-time photo-realistic portrait avatar from a single video. In: ACM SIGGRAPH 2023 Conference Proceedings (2023)

\bibitem{facevid2vid}
Wang, T.C., Mallya, A., Liu, M.Y.: One-shot free-view neural talking-head synthesis for video conferencing. In: Proceedings of the IEEE Conference on Computer Vision and Pattern Recognition (2021)

\bibitem{ssim}
Wang, Z., Bovik, A., Sheikh, H., Simoncelli, E.: Image quality assessment: from error visibility to structural similarity. IEEE Transactions on Image Processing  \textbf{13}(4),  600--612 (2004). \doi{10.1109/TIP.2003.819861}

\bibitem{x2face}
Wiles, O., Koepke, A.S., Zisserman, A.: X2face: A network for controlling face generation using images, audio, and pose codes. In: Proceedings of the European Conference on Computer Vision (ECCV) (September 2018)

\bibitem{latentavatar}
Xu, Y., Zhang, H., Wang, L., Zhao, X., Huang, H., Qi, G., Liu, Y.: Latentavatar: Learning latent expression code for expressive neural head avatar. In: Special Interest Group on Computer Graphics and Interactive Techniques Conference Conference Proceedings. SIGGRAPH ’23, ACM (Jul 2023). \doi{10.1145/3588432.3591545}, \url{http://dx.doi.org/10.1145/3588432.3591545}

\bibitem{learningnerf}
Yang, S., Wang, W., Lan, Y., Fan, X., Peng, B., Yang, L., Dong, J.: Learning dense correspondence for nerf-based face reenactment (2023)

\bibitem{styleheat}
Yin, F., Zhang, Y., Cun, X., Cao, M., Fan, Y., Wang, X., Bai, Q., Wu, B., Wang, J., Yang, Y.: Styleheat: One-shot high-resolution editable talking face generation via pre-trained stylegan. In: Computer Vision – ECCV 2022: 17th European Conference, Tel Aviv, Israel, October 23–27, 2022, Proceedings, Part XVII. p. 85–101. Springer-Verlag, Berlin, Heidelberg (2022). \doi{10.1007/978-3-031-19790-1_6}, \url{https://doi.org/10.1007/978-3-031-19790-1_6}

\bibitem{bilayer}
Zakharov, E., Ivakhnenko, A., Shysheya, A., Lempitsky, V.: Fast bi-layer neural synthesis of one-shot realistic head avatars. In: European Conference of Computer vision (ECCV) (August 2020)

\bibitem{zhao2021sparse}
Zhao, R., Wu, T., Guo, G.: Sparse to dense motion transfer for face image animation. In: 2021 IEEE/CVF International Conference on Computer Vision Workshops (ICCVW). pp. 1991--2000. IEEE Computer Society, Los Alamitos, CA, USA (oct 2021). \doi{10.1109/ICCVW54120.2021.00226}, \url{https://doi.ieeecomputersociety.org/10.1109/ICCVW54120.2021.00226}

\end{thebibliography}
\end{document}